\documentclass[runningheads]{llncs}

\usepackage{eccv}

\usepackage{eccvabbrv}

\usepackage{graphicx}
\usepackage{booktabs}
\usepackage{multirow}
\usepackage{wrapfig}
\usepackage{colortbl} 
\usepackage{tikz}
\usetikzlibrary{calc}
\usepackage{arydshln}

\usepackage[accsupp]{axessibility}  

\usepackage{hyperref}

\usepackage{orcidlink}

\usepackage{tabularray}
\usepackage{xcolor}
\usepackage[table]{xcolor}

\newcommand{\vlm}{\texttt{dinov3.txt}}
\usepackage{amssymb}
\usepackage{subcaption}
\usepackage{tikz}
\usepackage{arydshln}
\usepackage{pgfplots}
\usepgfplotslibrary{colorbrewer}
\pgfplotsset{compat=1.18}

\begin{document}

\title{\texttt{dinov3.seg}: Open-Vocabulary Semantic Segmentation with DINOv3} 

\titlerunning{\texttt{dinov3.seg}}

\author{Saikat Dutta\inst{1,2,3} \and
Biplab Banerjee\inst{2} \and
Hamid Rezatofighi\inst{3}}

\authorrunning{Dutta et al.}

\institute{IITB-Monash Research Academy \and
IIT Bombay \and
Monash University}

\maketitle

\begin{abstract}
Open-Vocabulary Semantic Segmentation (OVSS) assigns pixel-level labels from an open set of text-defined categories, demanding reliable generalization to unseen classes at inference. Although modern vision–language models (VLMs) support strong open-vocabulary recognition, their representations learned through global contrastive objectives remain suboptimal for dense prediction, prompting many OVSS methods to depend on limited adaptation or refinement of image–text similarity maps. This, in turn, restricts spatial precision and robustness in complex, cluttered scenes. We introduce \texttt{dinov3.seg}, extending \texttt{dinov3.txt} into a dedicated framework for OVSS. Our contributions are four-fold. First, we design a task-specific architecture tailored to this backbone, systematically adapting established design principles from prior open-vocabulary segmentation work.
Second, we jointly leverage text embeddings aligned with both the global \texttt{[CLS]} token and local patch-level visual features from ViT-based encoder, effectively combining semantic discrimination with fine-grained spatial locality. Third, unlike prior approaches that rely primarily on post hoc similarity refinement, we perform early refinement of visual representations prior to image–text interaction, followed by late refinement of the resulting image–text correlation features, enabling more accurate and robust dense predictions in cluttered scenes. Finally, we propose a high-resolution local–global inference strategy based on sliding-window aggregation, which preserves spatial detail while maintaining global context. 
We conduct extensive experiments on five widely adopted OVSS benchmarks to evaluate our approach. The results demonstrate its effectiveness and robustness, consistently outperforming current state-of-the-art methods.

\end{abstract}

\section{Introduction}
Open-Vocabulary Semantic Segmentation (OVSS) extends conventional semantic segmentation by allowing models to predict pixel-wise labels beyond a fixed, closed set of training categories. This flexibility is crucial for real-world deployments in robotics, autonomous driving, remote sensing, and medical imaging, where the label space is long-tailed, dynamic, and costly to annotate exhaustively. Despite rapid progress in vision--language models (VLMs) \cite{align,clip} for open-vocabulary recognition, translating their global image--text alignment into dense, fine-grained segmentation remains challenging. Representations learned primarily through global contrastive objectives often under-emphasize locality and spatial detail, which are essential for pixel-level discrimination---especially in cluttered scenes and high-resolution imagery.

\begin{figure}[!t]
\centering

\begin{tabular}{
c
@{\hspace{10pt}}
c
c
@{\hspace{10pt}} 
c
c
c
}

\multicolumn{1}{c}{} &
\multicolumn{2}{c}{\textbf{Pretrained models}} &
\multicolumn{3}{c}{\textbf{Fine-tuned models}} \\[2pt]

\begin{subfigure}{0.145\linewidth}\centering
\includegraphics[width=\linewidth]{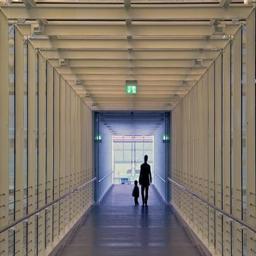}
\end{subfigure}
&
\begin{subfigure}{0.145\linewidth}\centering
\includegraphics[width=\linewidth]{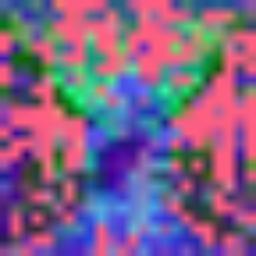}
\end{subfigure}
&
\begin{subfigure}{0.145\linewidth}\centering
\includegraphics[width=\linewidth]{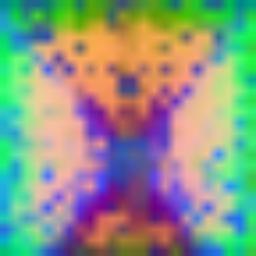}
\end{subfigure}
&
\begin{subfigure}{0.145\linewidth}\centering
\includegraphics[width=\linewidth]{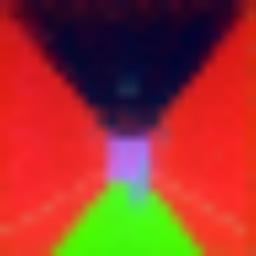}
\end{subfigure}
&
\begin{subfigure}{0.145\linewidth}\centering
\includegraphics[width=\linewidth]{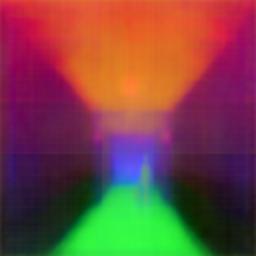}
\end{subfigure}
&
\begin{subfigure}{0.145\linewidth}\centering
\includegraphics[width=\linewidth]{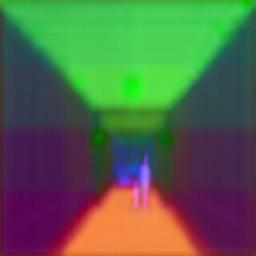}
\end{subfigure}

\\[-1pt]

\begin{subfigure}{0.145\linewidth}\centering\caption*{Input}\end{subfigure} &
\begin{subfigure}{0.145\linewidth}\centering\caption*{CLIP}\end{subfigure} &
\begin{subfigure}{0.145\linewidth}\centering\caption*{\texttt{dinov3.txt}}\end{subfigure} &
\begin{subfigure}{0.145\linewidth}\centering\caption*{\texttt{dinov3.txt} FT}\end{subfigure} &
\begin{subfigure}{0.145\linewidth}\centering\caption*{CAT-Seg}\end{subfigure} &
\begin{subfigure}{0.145\linewidth}\centering\caption*{Ours}\end{subfigure}

\end{tabular}

\vspace{-10px}

\caption{\textbf{Visualization of final visual features across different models.} Pretrained models exhibit noisy features, while \vlm\ produces comparatively cleaner features with more structural information than CLIP. Fine-tuned \vlm\ features yield sharper boundaries, but fail to capture fine-grained details. CLIP-based OVSS approach CAT-Seg recovers more detail in the features, yet produces less well-defined boundaries. In contrast, our method produces features with sharp boundaries and rich fine-grained detail.}
\label{fig:anyup_viz}
\vspace{-20px}
\end{figure}

Most OVSS pipelines \cite{san,liu2024scan,xie2024sed,catseg,maftp,peng2025hyperbolic,zhao2025dpseg,openbench} build on CLIP-style VLMs \cite{clip, siglip, evaclip} trained end-to-end via large-scale contrastive learning. While highly effective for image-level retrieval and classification, this training paradigm can bias features toward global semantics and weaken local cues, prompting downstream methods to rely on heuristic post-processing, prompt engineering, or late-stage refinement of similarity maps to recover spatial structure. In contrast, recently proposed \texttt{dino.txt} \cite{jose2025dinotxt} offers a complementary route. It builds on self-supervised DINO visual foundation models \cite{oquab2023dinov2,dinov3}, which are renowned for producing high-quality, spatially rich visual features. To equip these models with open-vocabulary capabilities, it employs locked-image text tuning (LiT) \cite{zhai2021lit}, aligning a text encoder to a frozen DINO backbone and thereby preserving the strong local structure learned through self-supervision. To support both global and dense tasks, it combines the \texttt{[CLS]} token with pooled patch features and introduces lightweight transformer adapters that bridge visual pre-training with image--text data, delivering state-of-the-art zero-shot performance across open-vocabulary global and dense tasks.

However, despite improved performance on open-vocabulary dense tasks compared to CLIP-style VLMs, \texttt{dino.txt} remains a VLM trained under a largely global contrastive objective and is predominantly evaluated in a zero-shot setting. 
Even when fine-tuned for segmentation, it fails to fully adapt to the open-vocabulary setting. As visualized in Fig.~\ref{fig:anyup_viz}, fine-tuned \texttt{dinov3.txt} features yield sharper boundaries but lose fine-grained detail, consistent with the limited performance on OVSS benchmarks observed in Sec.~\ref{subsec:ablations}.
Such adaptation lacks dedicated dense refinement, the utilization of complementary text representations for fine-grained vision-language correspondence and segmentation-aware optimization -- all critical for precise open-vocabulary pixel classification. This motivates a central research question: \textit{how can we explicitly optimize \texttt{dino.txt}-style VLMs for OVSS so that its dense features and image--text alignment become segmentation-aware, improving boundary fidelity and class separability while retaining open-vocabulary generalization?} 

We address this gap with \texttt{dinov3.seg}, the first OVSS framework explicitly built and trained on top of \texttt{dinov3.txt}, the DINOv3 instantiation of \texttt{dino.txt}. Our method is characterized by four synergistic design choices. First, we develop a task-specific segmentation architecture that moves beyond zero-shot reuse of \texttt{dinov3.txt}, drawing on common  practices from prior literature. Second, we model each class using complementary textual views aligned with both the global \texttt{[CLS]} token and local patch-level representations, and ensemble their image–text correlations to strengthen semantic discrimination while preserving spatial locality. Third, we introduce a dual-stage refinement scheme: an early refinement module that enhances dense visual representations before image–text interaction, improving feature discriminability, followed by a SAM-guided late refinement stage that denoises and regularizes image–text correlation maps for sharper boundaries and improved class consistency in cluttered scenes. Fourth, we employ a high-resolution local–global inference strategy based on sliding-window aggregation, enabling fine-grained spatial detail while maintaining global semantic coherence across large images. These components are mutually reinforcing: complementary global–local textual modeling yields enriched image–text correlations, early refinement improves feature quality prior to image-text interaction, late refinement stabilizes image-text correlation features, and high-resolution inference consolidates locally consistent predictions. Together, these advances form a principled extension of \texttt{dinov3.txt} that directly addresses localization and optimization bottlenecks in OVSS. We summarize our contributions as follows:

\vspace{-5px}

\begin{itemize}
\item Extension of \texttt{dinov3.txt} into a dedicated dense OVSS framework.

\item Integration of complementary global and local textual representations for stronger multimodal alignment.

\item Dual-stage refinement of visual features and image–text correlations.

\item High-resolution local–global inference via sliding-window aggregation.

\item State-of-the-art performance across five challenging OVSS benchmarks.
\end{itemize}

\vspace{-5px}

\vspace{-15px}

\section{Related Works}

\vspace{-5px}

\textbf{Open-Vocabulary Segmentation.}
OVSS assigns pixel-level labels from an open set of text-defined categories, and therefore requires simultaneously (i) reliable vision--language grounding and (ii) precise spatial localization for dense masks. Early \emph{pre-VLM} approaches aligned dense CNN features with fixed semantic spaces: ZS3Net~\cite{zs3net} synthesizes class-conditional features from Word2Vec guidance with DeepLab-v3+ supervision~\cite{deeplabv3p}, and SPNet~\cite{xian2019semantic} projects dense features into shared visual--semantic spaces for text-driven labeling. However, the cross-modal coupling is relatively weak in these models, often yielding brittle transfer to novel concepts and coarse object boundaries. With the rise of VLMs, CLIP-based pipelines such as LSeg~\cite{lseg}, OpenSeg~\cite{openseg}, and ZegFormer~\cite{zegformer,maskformer} considerably improve open-vocabulary recognition by strengthening image--text alignment. However, global contrastive pretraining in CLIP can bias representations toward image-level semantics and suppress fine-grained locality, compelling many methods to rely on post hoc refinement of similarity maps to recover spatial detail. Transformer-centric designs, including SAN~\cite{san}, FC-CLIP~\cite{fcclip}, and CAT-Seg~\cite{catseg}, integrate VLM cues through language-aware queries and attention biases, frozen convolutional backbones, or dense image--text cost volumes constructed and refined before decoding. Nevertheless, these designs remain susceptible to noisy cross-modal correspondences when the underlying alignment is not segmentation-aware, a limitation that becomes especially pronounced under clutter, occlusion, and high-resolution settings.
Diffusion-based methods (ODISE~\cite{odise}, DeDOS~\cite{dedos}, DP-Seg~\cite{zhao2025dpseg}) leverage rich intermediate semantics from generative models, typically at the expense of higher compute and greater sensitivity to feature extraction, prompting, or multi-stage processing. Segment Anything Model (SAM)-based hybrids (EB-Seg~\cite{ebseg}, ESC-Net~\cite{lee2025escnet}, USE~\cite{wang2024use}) incorporate strong mask priors or universal segment representations, yet their performance can be bounded by proposal quality and imperfect pixel-level alignment between segments and text labels. Recent efforts explore complementary directions such as parameter-efficient tuning, hyperbolic adaptation, multi-resolution training, or seen-class bias mitigation~\cite{peng2025peft,peng2025hyperbolic,zhu2024mrovseg,openbench}. 
Different from previous works, \texttt{dinov3.seg} upgrades the backbone signal itself by training an OVSS-specific architecture on top of \texttt{dinov3.txt}, combining textual semantic ensembling with early and late refinement schemes and an effective high-resolution local--global aggregation-based inference scheme, thereby improving boundary fidelity and spatial precision in complex scenes.

\textbf{DINO Family of Visual Foundation Models.}
DINO models learn visual representations via self-distillation, producing object-centric attention and spatially coherent features that transfer effectively to dense prediction tasks. DINOv2 scales this paradigm to larger data and stronger architectures, while DINOv3 further improves locality and correspondence under massive training and refined objectives, making it a particularly strong backbone for segmentation and dense matching. Building on these visual foundations, \texttt{dino.txt}~\cite{jose2025dinotxt} extends DINO to the vision--language regime via locked-image text tuning (LiT), aligning a text encoder to frozen DINO features to preserve locality while enabling open-vocabulary recognition and zero-shot dense prediction. Our work complements \texttt{dino.txt} by extending it into a fully trainable OVSS framework: \texttt{dinov3.seg} capitalizes on DINO's inherent spatial coherence through multi-granular image--text alignment and segmentation-aware optimization, yielding precise open-vocabulary masks that generalize robustly across seen and unseen classes.

\vspace{-10px}

\section{Methodology}

\vspace{-5px}

\subsection{Preliminaries: \texttt{dino.txt}}

\texttt{dino.txt} \cite{jose2025dinotxt} is a vision--language alignment framework built on top of a frozen DINOv2 visual backbone. It follows LiT strategy \cite{zhai2021lit}, where the visual encoder remains fixed and only the text encoder is trained, preserving the spatial structure and locality of self-supervised visual features. Image representations are constructed by combining the global \texttt{[CLS]} token with average-pooled patch features to support both image-level and dense prediction tasks. To bridge the domain gap between visual pre-training and image--text data, \texttt{dino.txt} introduces lightweight transformer blocks that align visual and textual embeddings using a contrastive objective. Although \texttt{dino.txt} is originally developed for DINOv2, the same text-alignment formulation is also available for DINOv3; hereafter, we refer to the DINOv2 and DINOv3 variants as \texttt{dinov2.txt} and \texttt{dinov3.txt}, respectively.

\begin{figure*}[!t]
    \centering
    \includegraphics[width=0.9\linewidth]{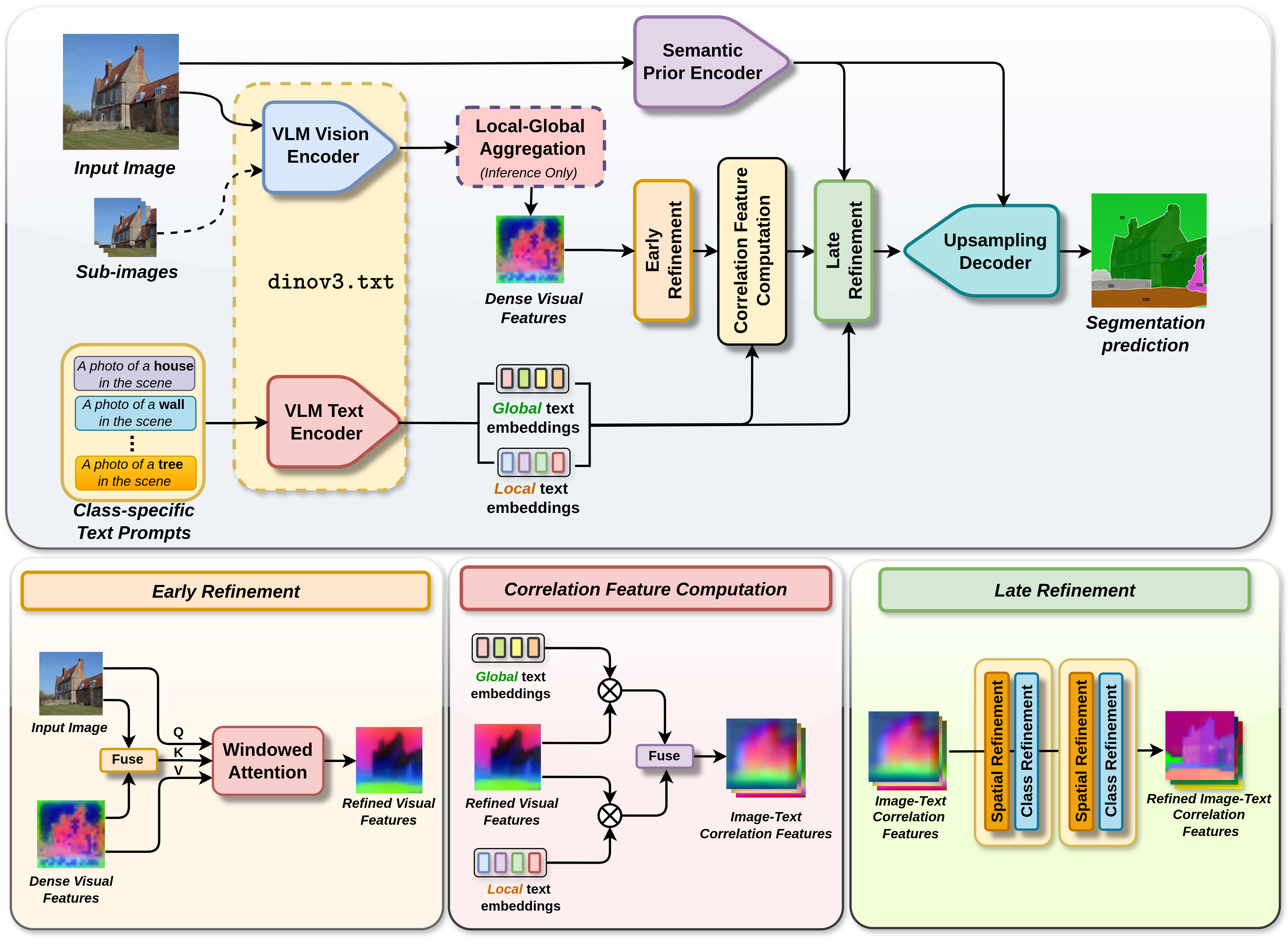}
    \caption{\textbf{Overall architecture of proposed \texttt{dinov3.seg}.} Given an input image (and sub-images at inference time), the \vlm\ backbone extracts dense visual features and multiple textual embeddings. The dense visual features are passed to the Early Refinement Module — via the Local-Global Aggregation module at inference time, or directly during training. The refined visual features interact with the textual embeddings to yield image–text correlation features, which are subsequently enriched with auxiliary semantic guidance from the Semantic Prior Encoder via the Late Refinement Module. The final segmentation map is generated by the Upsampling Decoder.}
    \label{fig:main}
    \vspace{-15px}
\end{figure*}

\subsection{Our proposal: \texttt{dinov3.seg}}

\textbf{Overview.} Given an input image $I$, the objective of Open-Vocabulary Semantic Segmentation is to assign a semantic label to each pixel from a set of categories defined by textual names or descriptions. Under the open-vocabulary setting, the class set available during training, denoted as $\mathcal{C}_{\text{train}}$, is not necessarily identical to the class set encountered at inference time, $\mathcal{C}_{\text{test}}$, thereby requiring the model to generalize beyond the seen categories. 

In this work, we propose a novel Open-Vocabulary Segmentation model, \texttt{dinov3.seg}, based on \vlm\ backbone. Our framework consists of the following components: (i) \textit{\vlm\ backbone} that extracts dense visual features from images and textual embeddings from category descriptions; (ii) an \textit{Early Refinement Module} that enhances the dense visual representations; (iii) a \textit{Semantic Prior Encoder} that provides auxiliary semantic guidance; (iv) a \textit{Correlation Feature Computation Module} that models image–text interactions to produce correlation features; (v) a \textit{Late Refinement Module} that further refines the correlation features using the semantic priors; and (vi) an \textit{Upsampling Decoder} that progressively decodes the refined features to generate the final segmentation output. We describe the model architecture in detail in the following paragraphs. An overview of the proposed framework is illustrated in Fig.~\ref{fig:main}.

\textbf{Visual and Textual Feature Extraction.}
Given an input image $I$, we extract dense visual features using the vision encoder of \vlm, denoted by $\mathcal{F}_v$:
\begin{equation}
    \phi_\text{CLS}, \phi_v = \mathcal{F}_v(I)
\end{equation}

where, $\phi_\text{CLS} \in \mathbb{R}^{C}$ represent \texttt{CLS} token and $ \phi_v \in \mathbb{R}^{C \times H \times W }$ represent final patch features from ViT-based vision encoder.

For each class $c \in \mathcal{C}_{\text{train}}$ (or $c \in \mathcal{C}_{\text{test}}$ during inference), we extract textual representations using the text encoder of \vlm, denoted by $\mathcal{F}_t$. Specifically, we construct a text prompt of the form ``A photo of a \textless class\textgreater\ in the scene'' and encode it as,
\begin{equation}
    \phi_t^{g}(c), \ \phi_t^{l}(c) = \mathcal{F}_t(c)
\end{equation}

where $\phi_t^{g}(c) \in \mathbb{R}^{C}$ is aligned with the $\phi_\text{CLS}$; we refer to this as the \emph{global text embedding}. $\phi_t^{l}(c) \in \mathbb{R}^{C}$ is aligned with the average of patch-wise visual features; we refer to this as the \emph{local text embedding}.

While \texttt{dinov3.txt} \cite{jose2025dinotxt} leverages only the local text embedding $\phi_t^{l}$ for open-vocabulary segmentation, we demonstrate that jointly utilizing both global and local text embeddings leads to more effective and robust segmentation performance in our framework (see Table~\ref{tab:config_ablation_updated}). The complementary nature of these embeddings is further validated in our analysis in the supplementary material.

\textbf{Semantic Prior Encoder (SPE).} 
In addition to the vision--language backbone, we introduce a semantic prior encoder $\mathcal{F}_\text{SPE}$ to provide complementary visual cues. Following \cite{dutta2025aeroseg}, we adopt the ViT-L based image encoder of Segment Anything Model (SAM)~\cite{sam} as our SPE. Experiments with alternative SPE choices are provided in the supplementary material.

We denote final-layer representation from $\mathcal{F}_\text{SPE}$ as $F_g^{(L)}$, which encodes high-level semantic context and global structural information for segmentation guidance.  Additionally, we also extract intermediate representations from the 7\textsuperscript{th} and 15\textsuperscript{th} transformer blocks, denoted as $F_g^{(7)}$ and $F_g^{(15)}$. Although all features share the same spatial resolution, they capture progressively richer abstractions, from structural cues to higher-level semantics. SPE is jointly fine-tuned with the rest of the architecture to ensure compatibility with the vision--language representations. Formally, given an input image $I$, semantic prior features are obtained as:

\begin{equation}
    F_g^{(7)}, F_g^{(15)}, F_g^{(L)} = \mathcal{F}_\text{SPE}(I).
\end{equation}

\textbf{Early Refinement of VLM features.}
Like most Vision--language models, \vlm\ is also optimized with a global contrastive objective, which often results in dense visual features $\phi_v$ that are noisy and lack the discriminative power required for fine-grained segmentation. To address this limitation, we introduce an early-stage refinement of visual features before any explicit image--text interaction. Crucially, this refinement must preserve the original image--text alignment so as not to disrupt compatibility with textual embeddings.

\begin{figure}[!t]
\centering
\setlength{\tabcolsep}{2pt}
\renewcommand{\arraystretch}{1.0}
\resizebox{\textwidth}{!}{
\begin{tabular}{ccc@{\hspace{12pt}}ccc}
\begin{tikzpicture}[inner sep=0]
  \node at (0,0) {\includegraphics[width=0.16\textwidth]{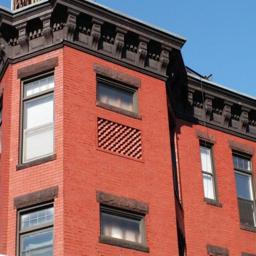}};
  \draw[blue, thick] (-0.5, 0.2) rectangle (0.5, 0.9);
\end{tikzpicture} &
\begin{tikzpicture}[inner sep=0]
  \node at (0,0) {\includegraphics[width=0.16\textwidth]{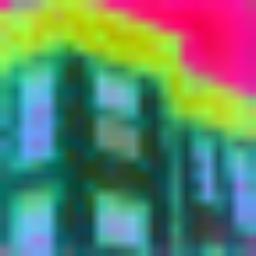}};
  \draw[blue, thick] (-0.5, 0.2) rectangle (0.5, 0.9);
\end{tikzpicture} &
\begin{tikzpicture}[inner sep=0]
  \node at (0,0) {\includegraphics[width=0.16\textwidth]{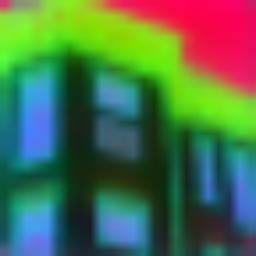}};
  \draw[blue, thick] (-0.5, 0.2) rectangle (0.5, 0.9);
\end{tikzpicture} &
\begin{tikzpicture}[inner sep=0]
  \node at (0,0) {\includegraphics[width=0.16\textwidth]{earlyref_viz/51/input.jpg}};
  \draw[cyan, thick] (0.1,-0.5) rectangle (0.7, 0.25);
\end{tikzpicture} &
\begin{tikzpicture}[inner sep=0]
  \node at (0,0) {\includegraphics[width=0.16\textwidth]{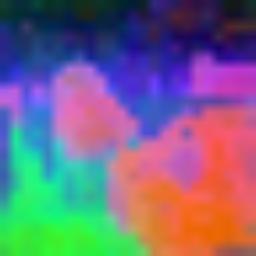}};
  \draw[cyan, thick] (0.1,-0.5) rectangle (0.7, 0.25);
\end{tikzpicture} &
\begin{tikzpicture}[inner sep=0]
  \node at (0,0) {\includegraphics[width=0.16\textwidth]{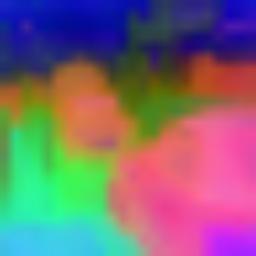}};
  \draw[cyan, thick] (0.1,-0.5) rectangle (0.7, 0.25);
\end{tikzpicture} \\
Input & w/o Early Ref. & w/ Early Ref. &
Input & w/o Early Ref. & w/ Early Ref. \\
\vspace{-15px}
\end{tabular}
}
\caption{Effect of Early Refinement on \vlm\ visual features.}
\vspace{-15px}
\label{fig:earlyref}
\end{figure}
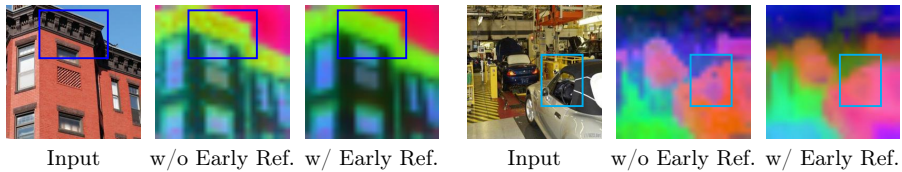

We adopt the transformer-based AnyUp~\cite{wimmer2025anyup} module to refine $\phi_v$. Although originally designed for guided feature upsampling, AnyUp can be fine-tuned to effectively refine dense visual representations for open-vocabulary segmentation. Visualization of its effect on \vlm\ features is shown in Fig~\ref{fig:earlyref}.  

Concretely, given an input image $I$ and dense visual features $\phi_v$, the image is first processed by a lightweight convolutional encoder to obtain intermediate features $\psi_v$. Rotary positional encodings (RoPE) \cite{rope} are added to $\psi_v$ to preserve spatial structure. Refinement is performed via a window-based attention module~\cite{ramachandran2019stand}, where queries are computed from $\psi_v$, keys from both $\psi_v$ and $\phi_v$, and values are the original dense features $\phi_v$. Since values are drawn solely from $\phi_v$, the output is a linear recombination of original patch-level representations, ensuring that the refined features $\phi_v$ remain anchored to the \vlm\ feature space and retain image–text alignment: $\phi_v^\text{ref} = \mathrm{EarlyRef}(\phi_v).$

\textbf{Image--Text Correlation Feature Computation.}
To explicitly model image--text correspondence, we compute cosine similarity between dense visual features and textual embeddings for each spatial location and each semantic class. Given refined dense visual features $\phi_v^\text{ref} \in \mathbb{R}^{C \times H \times W}$ and class-wise textual embeddings, we compute for every spatial location $(h,w)$ and class $c$:
\vspace{-5px}
\begin{equation}
\begin{aligned}
s^{g}(c,h,w) &= \cos\!\left(\phi_v^{\text{ref}}(h,w), \phi_{t}^{g}(c)\right), \\
s^{l}(c,h,w) &= \cos\!\left(\phi_v^{\text{ref}}(h,w), \phi_{t}^{l}(c)\right).
\end{aligned}
\label{eq:similarities}
\end{equation}
\vspace{-5px}

where $\phi_{t}^{g}(c)$ and $\phi_{t}^{l}(c)$ denote the global and local textual embeddings of class $c$. The resulting similarity tensors $S^{g}, S^{l} \in \mathbb{R}^{N \times H \times W}$ encode class-wise semantic alignment at each spatial location, with $N$ denoting the number of classes.

We concatenate the similarity tensors along the channel dimension and project them using a convolutional layer to obtain higher-level correlation features:
\begin{equation}
\phi_{\text{corr}} = \mathrm{Conv}\!\left([S^{g}; S^{l}]\right),
\quad
\phi_{\text{corr}} \in \mathbb{R}^{N \times C_{\text{corr}} \times H \times W}.
\label{eq:corr_feature}
\end{equation}
where $C_{\text{corr}}$ is the number of output channels. We refer to $\phi_{\text{corr}}$ as the \emph{image--text correlation features}. These features are subsequently refined by Late Refinement module to improve semantic consistency and localization accuracy.

\textbf{Late Refinement of Correlation Features.}
In addition to refining dense visual features from the vision--language model, we further refine the image--text correlation features in a late refinement stage. While early visual refinement improves feature quality prior to image--text interaction, refining the correlation features themselves is crucial for producing spatially accurate and class-consistent segmentation outputs. Following \cite{catseg,dutta2025aeroseg}, our late refinement module consists of two complementary components - \textit{Spatial Refinement Block} and \textit{Class Refinement Block}.

\textbf{\textit{Spatial Refinement Block:}}
To enhance the spatial coherence of the class-wise correlation features, we employ a Swin Transformer~\cite{liu2021swin}-based refinement module. Refinement is performed independently for each class using two successive Swin Transformer blocks: the first applies Window Multi-head Self-Attention (W-MSA) to model local spatial interactions, followed by a Shifted Window Multi-head Self-Attention (SW-MSA) block to enable cross-window information exchange.

To further strengthen spatial modeling, we incorporate auxiliary guidance from SPE into the Spatial Refinement Block. Specifically, SPE feature $F_g^{(L)}$ is projected into the correlation feature space and used to guide spatial refinement. For each class $c$, the refined correlation feature $\phi'_\text{corr}(:,:,c)$ is computed as,
\begin{equation}
    \phi'_\text{corr}(:,:,c) = \mathrm{SpatialRef}\!\left(\phi_\text{corr}(:,:,c),\, \mathcal{M}_v\!\left(F_g^{(L)}\right)\right),
    \label{eq:spatial_refinement}
\end{equation}
where $\mathrm{SpatialRef}(\cdot)$ denotes the Spatial Refinement Block, $\phi_\text{corr}(:,:,c)$ represents the correlation feature corresponding to class $c$, and $\mathcal{M}_v(\cdot)$ is an MLP projection layer applied to the SPE feature.

\textbf{\textit{Class Refinement Block.}}
While the Spatial Refinement Block focuses on improving spatial coherence, the Class Refinement Block aims to enhance class-wise discrimination at each spatial location. We refine the correlation features by explicitly modeling semantic relationships across classes.

Specifically, we apply a transformer-based refinement module along the class dimension, where attention is computed across class channels for each spatial location. This allows the model to capture inter-class dependencies and suppress ambiguous class predictions. For each spatial location $(h,w)$, the refined correlation feature $\phi^{\prime\prime}_{\mathrm{corr}}(h,w,:)$ is computed as,
\begin{equation}
    \phi^{\prime\prime}_{\mathrm{corr}}(h,w,:)
    =
    \mathrm{ClassRef}\!\left(
        \phi^{\prime}_{\mathrm{corr}}(h,w,:), \mathcal{M}_t(\bar{\phi}_t)
    \right),
    \label{eq:class_refinement}
\end{equation}
where $\mathrm{ClassRef}(\cdot)$ denotes the Class Refinement Block, and $\phi^{\prime}_{\mathrm{corr}}(h,w,:)$ represents the class-wise correlation responses at spatial location $(h,w)$. $\mathcal{M}_t(\cdot)$ is an MLP projection layer applied to the textual guidance feature, $\bar{\phi}_t = (\phi_t^g + \phi_t^l)/2$.

In our framework, the combination of $\mathrm{SpatialRef}$ and $\mathrm{ClassRef}$ is applied twice in succession, enabling progressive refinement of correlation features by jointly improving spatial coherence and inter-class discrimination.

\textbf{Upsampling Decoder.}
Since the refined correlation features $\phi^{\prime\prime}_{\mathrm{corr}}$ are at $\tfrac{1}{16}$ of the input image resolution, we employ a lightweight convolutional upsampling decoder to progressively recover full spatial resolution. First, $\phi^{\prime\prime}_{\mathrm{corr}}$ is upsampled by a factor of 2 using a transposed convolution, yielding an intermediate feature map $\phi^{\prime\,2\times}_{\mathrm{corr}}$. The upsampled correlation feature $\phi^{2\times}_{\mathrm{corr}}$ is obtained by concatenating $\phi^{\prime\,2\times}_{\mathrm{corr}}$  with upsampled SPE guidance features and applying a convolutional fusion:
\begin{equation}
    \phi^{2\times}_{\mathrm{corr}}
    =
    \operatorname{conv}\!\left(
        \bigl[\phi^{\prime\,2\times}_{\mathrm{corr}},\, 
        \operatorname{up}(F_g^{(7)},2)\bigr]
    \right).
    \label{eq:upsampled_feature}
\end{equation}
This procedure is repeated with SPE guidance feature $F_g^{(15)}$ 
to yield $\phi^{4\times}_{\mathrm{corr}}$, followed by a final 
convolution layer and bilinear upsampling to produce the full-resolution 
segmentation prediction $\hat{y}$.

\begin{figure}[!t]
    \centering
    \includegraphics[width=0.6\linewidth]{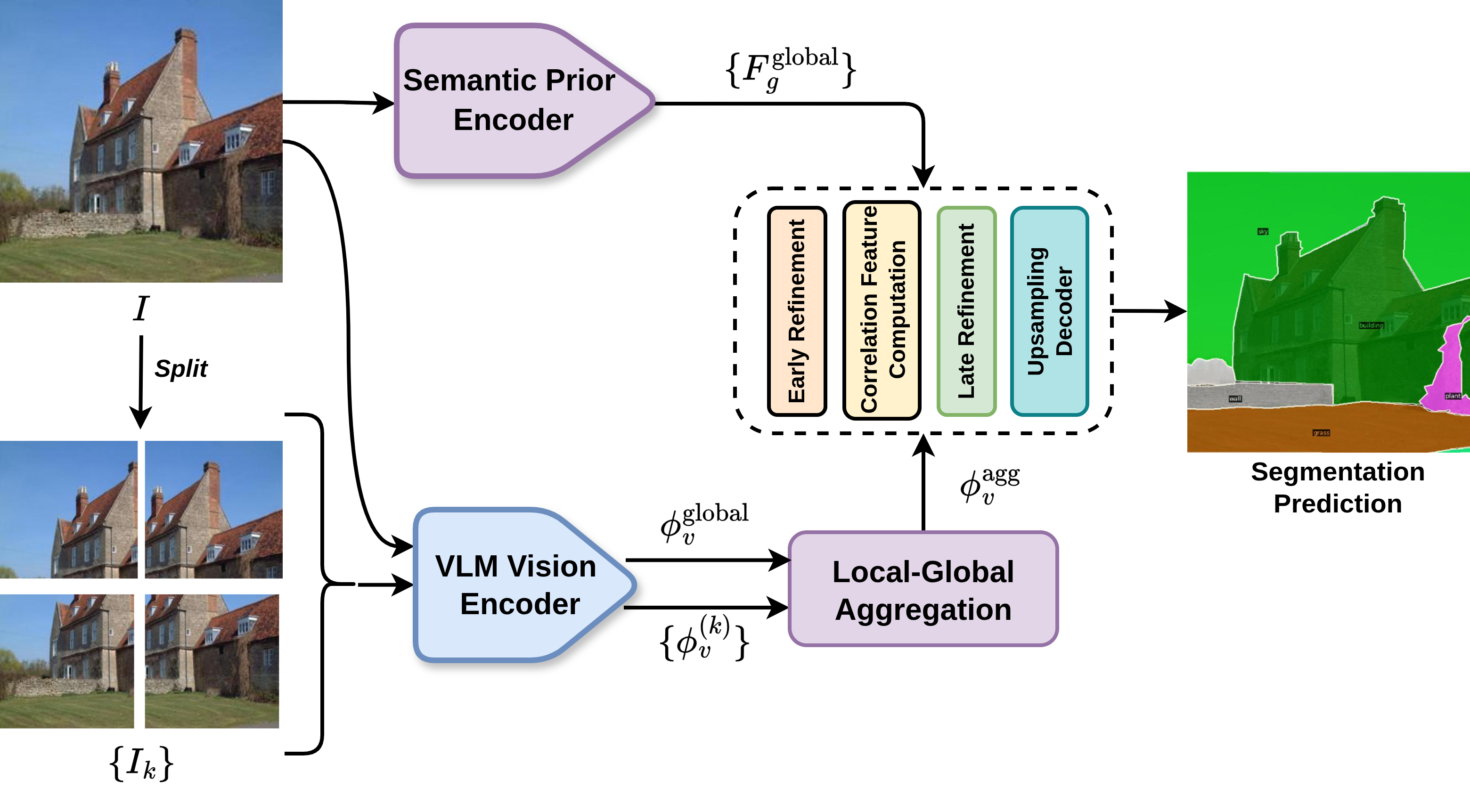}
    \vspace{-10px}
    \caption{Our proposed Local-Global Aggregation (LGA) Inference Strategy.}
    \label{fig:inf}
    \vspace{-15px}
\end{figure}

\textbf{Inference via Local--Global Aggregation (LGA).} During training, our model operates on image crops of size $384 \times 384$. To enable inference on high-resolution images, we design a sliding-window-based strategy inspired by ~\cite{trident, catseg}. Specifically, the input image $I$ is first resized to $640 \times 640$ and partitioned into overlapping sub-images $\{I_k\}$ of size $384 \times 384$, with an overlap of $128 \times 128$ pixels. Each sub-image is processed independently by the \vlm\ vision encoder. In parallel, the resized full image is encoded to obtain global VLM features:
\begin{equation}
\phi_v^{(k)} = \mathcal{F}_v(I_k), \quad \phi_v^{\text{global}} = \mathcal{F}_v(I).
\end{equation}

The resulting sub-image features are merged by accounting for overlapping regions, producing locally aggregated VLM features, denoted as $\bar{\phi}_v$. The local and global VLM features are then fused via simple averaging:
\begin{equation}
    \phi_v^{\text{agg}} = \tfrac{1}{2}\left(\bar{\phi}_v + \phi_v^{\text{global}}\right).
\end{equation}
We refer to this strategy as \emph{Local--Global Aggregation (LGA)}. For the SPE, we directly operate on resized full image to extract guidance features:
\begin{equation}
    \{F_g^{\text{global}}\} = \mathcal{F}_\text{SPE}(I).
\end{equation}

Finally, the aggregated VLM features $\phi_v^{\text{agg}}$ and the guidance features $\{F_g^{\text{global}}\}$ are fed into subsequent modules to produce the final segmentation map. Overall schematic of the proposed inference strategy is shown in Fig.~\ref{fig:inf}. We validate the inference strategy design choices via an ablation study in Sec.~\ref{subsec:ablations}.

\vspace{-10px}

\subsection{Loss functions}
The following loss functions are used to train our model. 

\textbf{Focal loss} \cite{lin2017focal} is a variant of Binary Cross-Entropy loss that addresses 
the imbalance between easy and hard samples, a prevalent challenge in segmentation. Focal loss is given by,
\begin{equation}
\mathcal{L}_{\text{focal}}(\hat{y}, y)
= -\tfrac{1}{HW} \sum_{i=1}^{HW}
\left[
(1 - \hat{y}_i)^{\gamma} y_i \log(\hat{y}_i)
+ \hat{y}_i^{\gamma} (1 - y_i) \log(1 - \hat{y}_i)
\right]
\end{equation}

where $\hat{y}$ is the predicted probability, $y$ is the ground truth label, and $\gamma$ is the focusing parameter that controls the down-weighting of easy samples.

\textbf{Dice loss} \cite{dice_loss} maximizes the Intersection over Union (IoU) between predicted map and ground truth segmentation mask. Dice loss is given by, 

\begin{equation}
\mathcal{L}_{\text{dice}}(\hat{y}, y)
= 1 - \tfrac{2 \sum_{i=1}^{HW} y_i \hat{y}_i}
{\sum_{i=1}^{HW} y_i + \sum_{i=1}^{HW} \hat{y}_i}
\end{equation}

We adopt a weighted combination of Focal loss and Dice loss~\cite{cheng2021per, zhou2022zegclip}, 
given by $\mathcal{L} = \mathcal{L}_\text{focal} + \lambda \mathcal{L}_\text{dice}$. 
Our experiments in Sec.~\ref{subsec:ablations} demonstrate that this combination leads to more 
effective segmentation performance over the conventional Binary Cross-Entropy 
loss~\cite{catseg,xie2024sed,zhao2025dpseg,peng2025hyperbolic}.

\vspace{-10px}
\section{Experiments}

\vspace{-5px}

\subsection{Dataset Description}

We train our method on the COCO-Stuff \cite{caesar2018coco} train split consisting of 118K images, and evaluate on ADE20K \cite{zhou2019semantic}, Pascal Context \cite{mottaghi2014role}, and Pascal VOC \cite{everingham2009pascal}. ADE20K is a diverse indoor–outdoor benchmark with 2K validation images, and we report results under both the 150-class (A-150) and 847-class (A-847) settings. Pascal Context contains 5K validation images covering a wide range of scenes, and we evaluate on both the 59-class (PC-59) and 459-class (PC-459) variants. Pascal VOC (PAS-20) includes 1.5K validation images with 20 classes.

\vspace{-5px}

\subsection{Implementation details}

We implement our method using PyTorch and the Detectron2 framework. The vision--language backbone, the semantic prior encoder and Early Refinement module are finetuned with an initial learning rate of $2 \times 10^{-6}$, while the remaining modules use a higher initial learning rate of $2 \times 10^{-4}$. All models are optimized using AdamW with a batch size of 4. We employ a cosine learning rate schedule and train the models for 80K iterations.
The values of $\lambda$ and 
$\gamma$ are set to $0.05$ and $2$ empirically. Performance under varying $\lambda$ and $\gamma$ values is reported in Supplementary Material.
All experiments are conducted on a server equipped with four NVIDIA A100 80 GB GPUs. We evaluate our models using the mean Intersection-over-Union (mIoU) metric.


\begin{table*}[!t]
\centering
\caption{\textbf{Quantitative comparison with state-of-the-art OVSS methods.}
Best and second-best results are highlighted in \textcolor{red}{red} and \textcolor{blue}{blue}, respectively.
Avg denotes the mean mIoU over A-847, PC-459, A-150, PC-59, and PAS-20.}
\label{tab:ovss_sota}
\resizebox{0.9\textwidth}{!}{
\begin{tblr}{
  colspec = {l c c c c c c|},
  row{1}  = {font=\bfseries},      
  hline{1,2} = {-}{0.08em},
  row{18} = {bg=gray!20},
  rowsep = 2pt,
  stretch = 0.9
}
Method & Venue &
A-847 & PC-459 & A-150 & PC-59 & PAS-20 & Avg. \\

OVSeg \cite{ovseg} & CVPR'23 & 9.0 & 12.4 & 29.6 & 55.7 & 94.5 & 40.24 \\ 
SAN \cite{san}       & CVPR'23     & 12.4 & 15.7 & 32.1 & 57.7 & 94.6 & 42.50 \\
ODISE \cite{odise}         & CVPR'23     & 11.1 & 14.5 & 29.9 & 57.3 & --   & -- \\
FC-CLIP \cite{fcclip}      & NeurIPS'23  & 14.8 & 18.2 & 34.1 & 58.4 & 95.4 & 44.18 \\


SCAN \cite{liu2024scan}     & CVPR'24     & 14.0 & 16.7 & 33.5 & 59.3 & 97.2 & 44.14 \\
SED \cite{xie2024sed}       & CVPR'24     & 13.9 & 22.6 & 35.2 & 60.6 & 96.1 & 45.68 \\
CAT-Seg \cite{catseg}        & CVPR'24     & 16.0 & 23.8 & 37.9 & 63.3 & 97.0 & 47.60 \\
MAFT+ \cite{maftp} & ECCV'24 & 15.1 & 21.6 &  36.1 & 59.4 & 96.5 & 45.74 \\ 

EOV-Seg \cite{eovseg}        & AAAI'25     & 12.8 & 16.8 & 32.1 & 56.9 & 94.8 & 42.68 \\
H-CLIP \cite{peng2025peft}   & CVPR'25     & 16.5 & 24.2 & 38.4 & 64.1 & 97.7 & 48.18 \\
HyperCLIP \cite{peng2025hyperbolic}
                            & CVPR'25     & 16.3 & 24.1 & 38.2 & 64.2 & \textcolor{blue}{98.3} & 48.22 \\
ESCNet \cite{lee2025escnet}  & CVPR'25     & \textcolor{blue}{18.1} & \textcolor{blue}{27.0} & \textcolor{blue}{41.8} & \textcolor{blue}{65.6} & \textcolor{blue}{98.3} & \textcolor{blue}{50.16} \\
DPSeg \cite{zhao2025dpseg}   & CVPR'25     & 15.7 & 24.1 & 37.1 & 62.3 & \textcolor{red}{98.5} & 47.54 \\
DEDOS \cite{dedos}           & ICCV'25     & 17.9 & 25.6 & 39.4 & \textcolor{red}{65.7} & 97.6 & 49.24 \\
OVSNet \cite{openbench}      & ICCV'25     & 16.2 & 23.5 & 37.1 & 62.0 & 96.9 & 47.14 \\
SAM3 \cite{sam3} & ICLR'26 & 13.8 & 18.8 & 39.0 & 60.8 & - & - \\    

 \textbf{\texttt{dinov3.seg} (Ours)} & -- &
\textcolor{red}{20.09} & \textcolor{red}{27.80} & \textcolor{red}{42.19} & 64.27 & 97.86 & \textcolor{red}{50.44} \\
\hline
\end{tblr}
}
\vspace{-10px}
\end{table*}

\subsection{Comparison with state-of-the-art}

We compare \texttt{dinov3.seg} against a broad set of state-of-the-art open-vocabulary semantic segmentation methods, including OVSeg~\cite{ovseg}, SAN~\cite{san}, ODISE~\cite{odise}, FC-CLIP~\cite{fcclip}, SCAN~\cite{liu2024scan}, SED~\cite{xie2024sed}, CAT-Seg~\cite{catseg}, MAFT+~\cite{maftp}, EOV-Seg~\cite{eovseg}, H-CLIP~\cite{peng2025peft}, HyperCLIP~\cite{peng2025hyperbolic}, ESCNet~\cite{lee2025escnet}, DPSeg~\cite{zhao2025dpseg}, DEDOS~\cite{dedos}, and OVSNet~\cite{openbench}. Additionally, we compare against SAM3~\cite{sam3}, a unified foundational segmentation model. For a fair comparison, we report results using the \emph{large} variants of all competing models whenever available. Although \texttt{dinov3.txt} serves as our primary VLM backbone, our framework generalizes across alternative backbones with minimal adjustments while maintaining consistently strong results; full details are provided in the supplementary material.

Table~\ref{tab:ovss_sota} reports quantitative results across benchmarks. \texttt{dinov3.seg} attains the best average mIoU across all datasets. 
In particular, it achieves the top performance on A-847, PC-459, and A-150, while remaining competitive on PC-59 and PAS-20. The substantial gains on A-847, PC-459, and A-150 highlight strong scalability to large and fine-grained category sets that demand robust generalization to numerous unseen classes. In contrast, PC-59 and PAS-20 exhibit significant overlap with the COCO-Stuff training categories, as noted in prior work~\cite{san} and further supported by our seen--unseen class partition (details in the supplementary material). Only 3 out of 59 classes in PC-59 and
\emph{none} of the 20 classes in PAS-20 qualify as unseen. Consequently, results on these benchmarks largely reflect seen-class optimization rather than open-vocabulary ability, and our method
remains competitive even under these conditions.

To further examine generalization, Table~\ref{tab:seen_unseen} reports mIoU on seen and unseen class subsets using the same similarity-based partition. \texttt{dinov3.seg} consistently improves both seen and unseen performance over state-of-the-art methods across all datasets. Notably, on PC-59 and PC-459, the gains on unseen classes are substantially larger than those on seen classes, indicating that the proposed design primarily enhances open-vocabulary generalization and recognition of novel categories.

\begin{table}[!t]

\centering
\caption{\textbf{Seen and unseen class mIoU comparison across datasets.} PAS-20 is omitted as all its categories are classified as seen under our similarity-based partition. $\Delta$ indicates absolute difference between our method and CAT-Seg.}
\renewcommand{\arraystretch}{0.9}

\resizebox{\linewidth}{!}{%
\begin{tabular}{
l
cc @{\hspace{10pt}}
cc @{\hspace{10pt}}
cc @{\hspace{10pt}}
cc
}
\hline
 & \multicolumn{2}{c}{\textbf{A-847}} 
 & \multicolumn{2}{c}{\textbf{PC-459}} 
 & \multicolumn{2}{c}{\textbf{A-150}}
 & \multicolumn{2}{c}{\textbf{PC-59}} \\
Method 
 & Seen & Unseen 
 & Seen & Unseen 
 & Seen & Unseen
 & Seen & Unseen \\
\hline
SED \cite{xie2024sed}
 & 31.67 & 12.23
 & 40.22 & 15.86
 & 47.02 & 28.07
 & 61.73 & 40.35 \\
 MAFT+ \cite{maftp}
 & 38.11 & 13.40
 & 39.71 & 14.01
 & 48.47 & 29.11
 & 61.11 & 29.37 \\
CAT-Seg \cite{catseg}
 & 38.67 & 13.80 
 & 42.83 & 16.40
 & 48.92 & 31.17 
 & 64.68 & 37.90 \\
\texttt{dinov3.seg} (Ours) 
 & \textbf{43.07} & \textbf{17.99} 
 & \textbf{44.72} & \textbf{21.27}
 & \textbf{54.06} & \textbf{35.11} 
 & \textbf{65.43} & \textbf{42.43} \\
\hline
$\Delta$
 & +4.39 & +4.19
 & +1.89 & +4.87
 & +5.14 & +3.95
 & +0.75 & +4.53 \\
\hline
\end{tabular}%
}
\vspace{-15px}
\label{tab:seen_unseen}
\end{table}

\begin{figure*}[!h]
\centering
\setlength{\tabcolsep}{2pt}
\renewcommand{\arraystretch}{1.0}
\begin{tabular}{ccccc}

\begin{tikzpicture}[inner sep=0]
\node at (0,0){\includegraphics[width=0.18\textwidth]{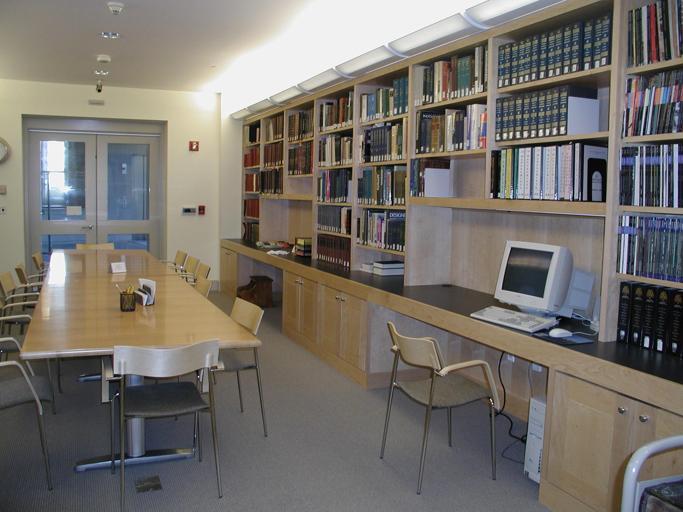}};
\draw[red, thick, densely dashed] (0,0) rectangle (1.0,0.8);
\draw[blue, thick, densely dashed] (0.1,-0.8) rectangle (0.75,-0.2);
\end{tikzpicture} &

\begin{tikzpicture}[inner sep=0]
\node at (0,0){\includegraphics[width=0.18\textwidth]{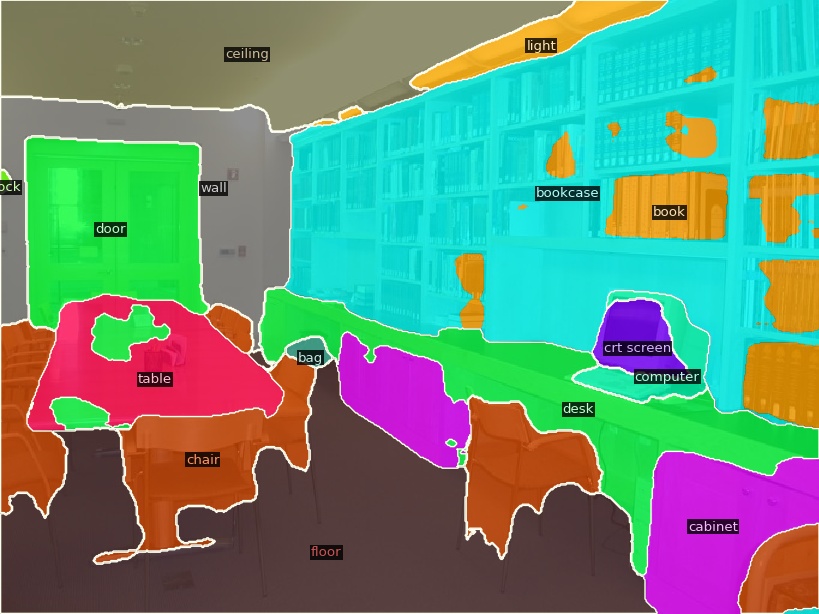}};
\draw[red, thick, densely dashed] (0,0) rectangle (1.0,0.8);
\draw[blue, thick, densely dashed] (0.1,-0.8) rectangle (0.75,-0.2);
\end{tikzpicture} &

\begin{tikzpicture}[inner sep=0]
\node at (0,0){\includegraphics[width=0.18\textwidth]{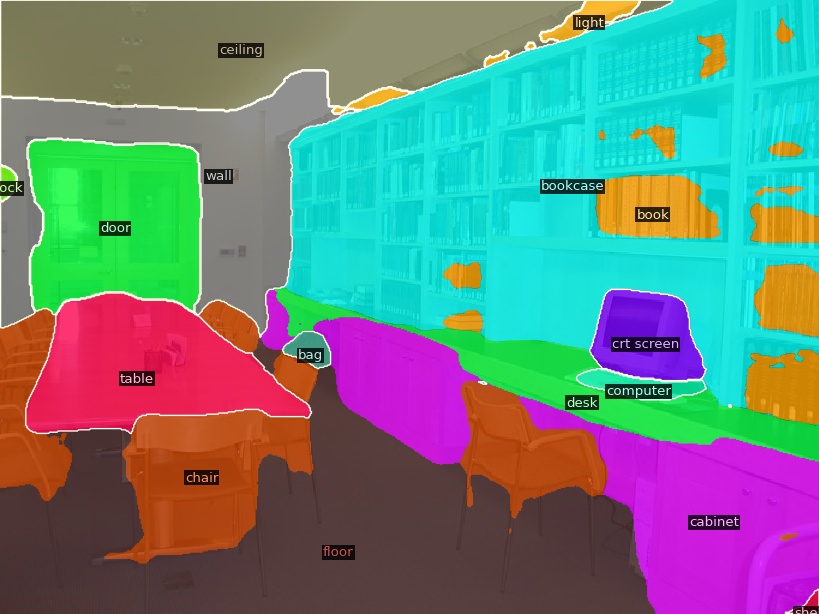}};
\draw[red, thick, densely dashed] (0,0) rectangle (1.0,0.8);
\draw[blue, thick, densely dashed] (0.1,-0.8) rectangle (0.75,-0.2);
\end{tikzpicture} &

\begin{tikzpicture}[inner sep=0]
\node at (0,0){\includegraphics[width=0.18\textwidth]{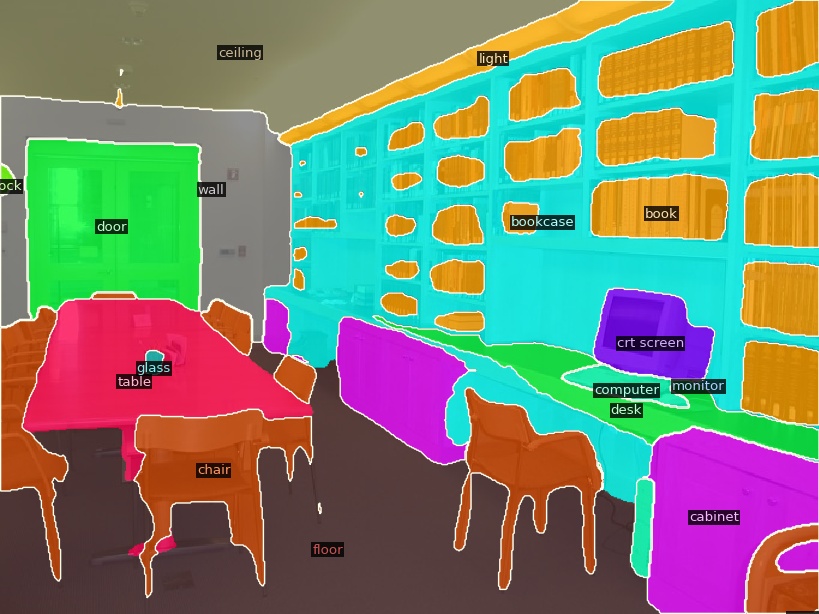}};
\draw[red, thick, densely dashed] (0,0) rectangle (1.0,0.8);
\draw[blue, thick, densely dashed] (0.1,-0.8) rectangle (0.75,-0.2);
\end{tikzpicture} &

\begin{tikzpicture}[inner sep=0]
\node at (0,0){\includegraphics[width=0.18\textwidth]{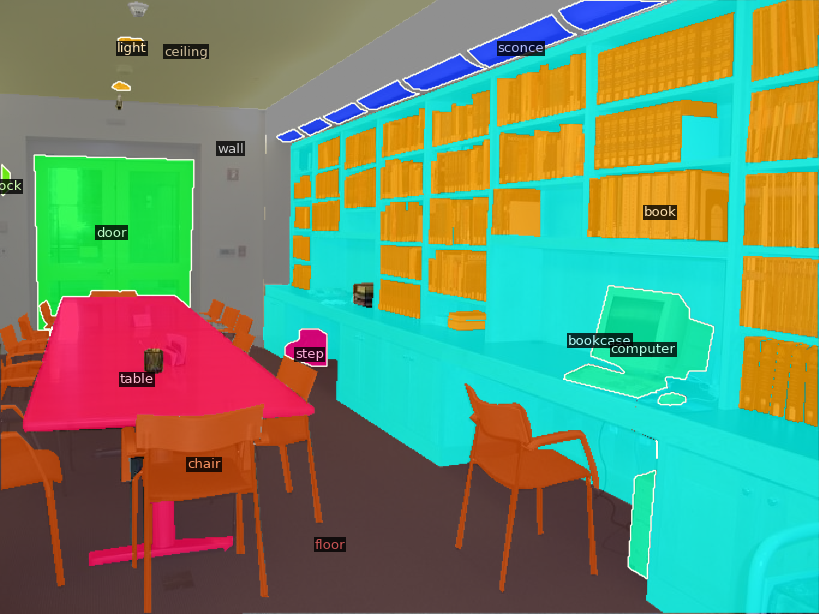}};
\draw[red, thick, densely dashed] (0,0) rectangle (1.0,0.8);
\draw[blue, thick, densely dashed] (0.1,-0.8) rectangle (0.75,-0.2);
\end{tikzpicture} \\

\begin{tikzpicture}[inner sep=0]
\node at (0,0){\includegraphics[width=0.18\textwidth]{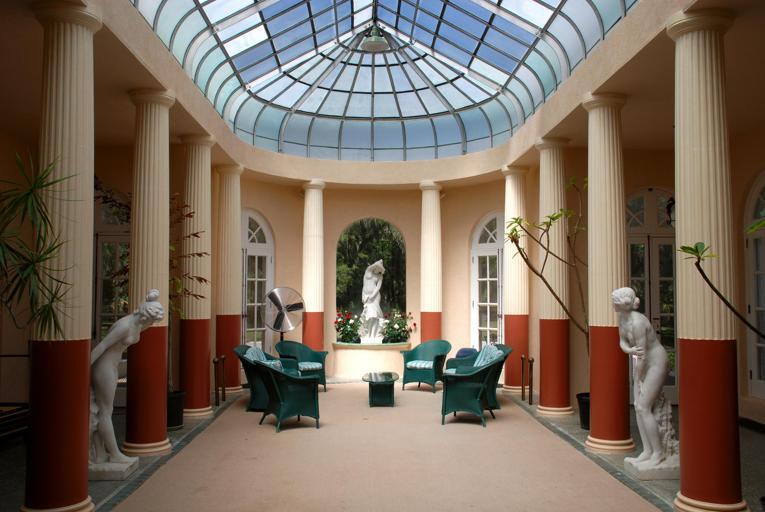}};
\draw[blue, thick, densely dashed] (0.25,-0.4) rectangle (0.8,0.75);
\end{tikzpicture} &

\begin{tikzpicture}[inner sep=0]
\node at (0,0){\includegraphics[width=0.18\textwidth]{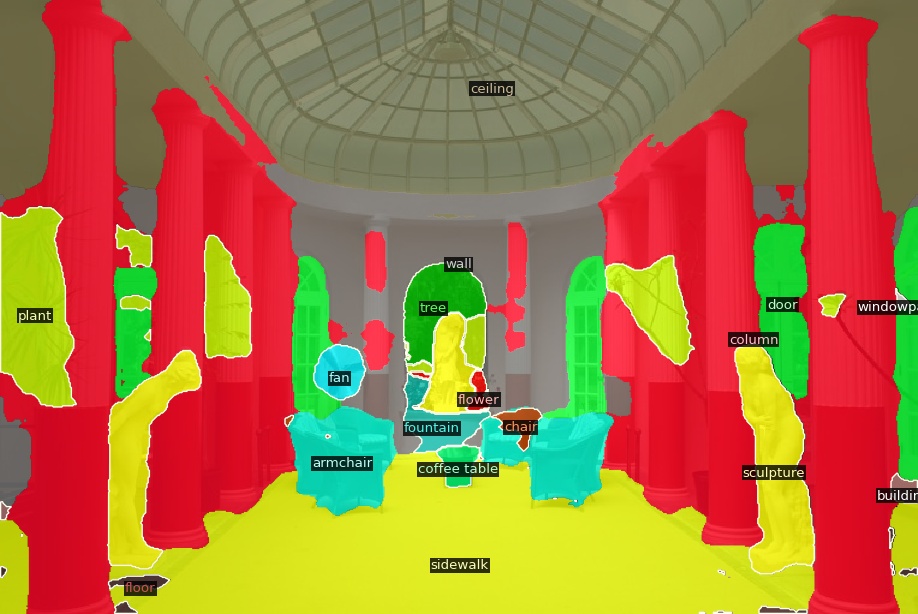}};
\draw[blue, thick, densely dashed] (0.25,-0.4) rectangle (0.8,0.75);
\end{tikzpicture} &

\begin{tikzpicture}[inner sep=0]
\node at (0,0){\includegraphics[width=0.18\textwidth]{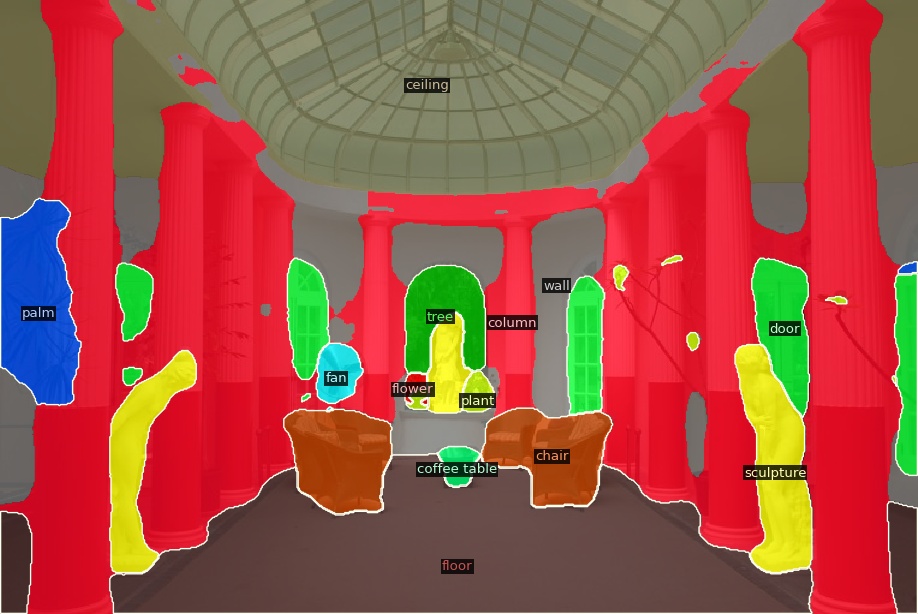}};
\draw[blue, thick, densely dashed] (0.25,-0.4) rectangle (0.8,0.75);
\end{tikzpicture} &

\begin{tikzpicture}[inner sep=0]
\node at (0,0){\includegraphics[width=0.18\textwidth]{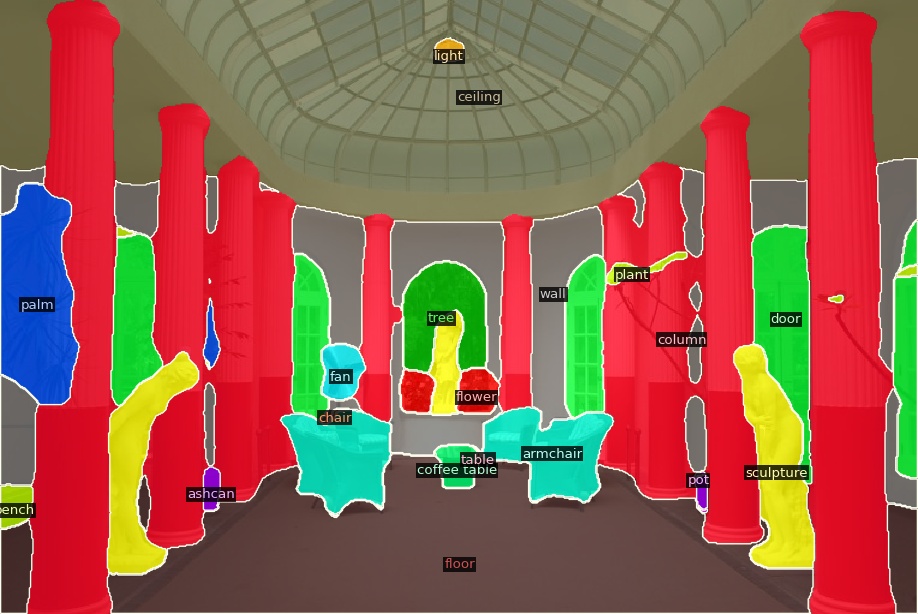}};
\draw[blue, thick, densely dashed] (0.25,-0.4) rectangle (0.8,0.75);
\end{tikzpicture} &

\begin{tikzpicture}[inner sep=0]
\node at (0,0){\includegraphics[width=0.18\textwidth]{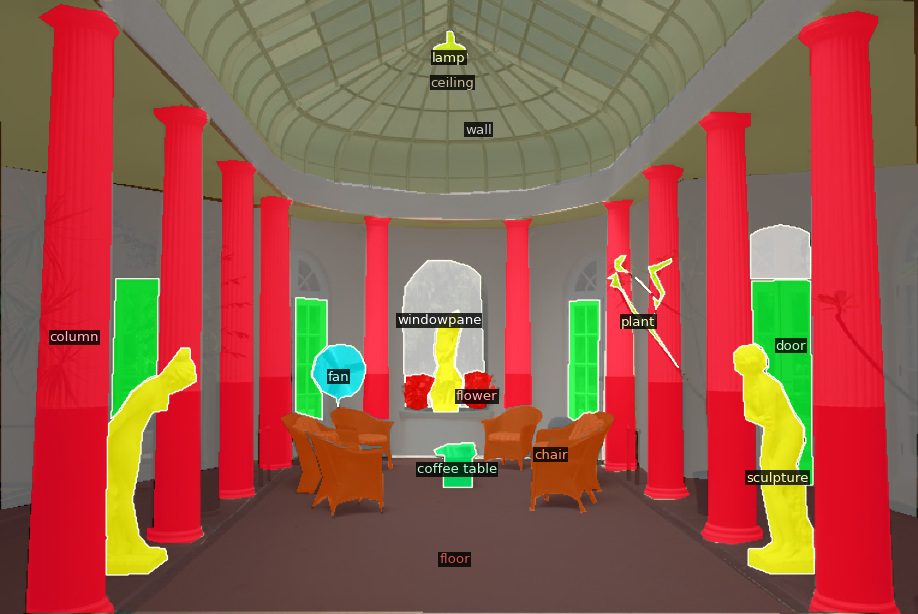}};
\draw[blue, thick, densely dashed] (0.25,-0.4) rectangle (0.8,0.75);
\end{tikzpicture} \\

\begin{tikzpicture}[inner sep=0]
\node at (0,0){\includegraphics[width=0.18\textwidth]{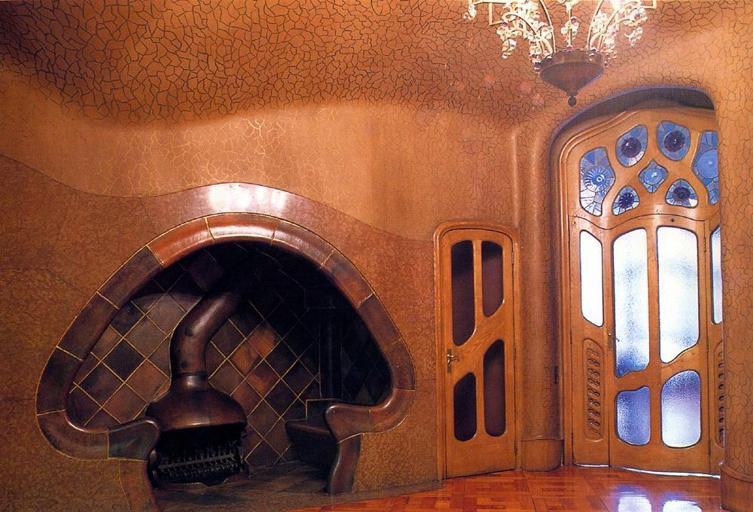}};
\draw[blue, thick, densely dashed] (-0.9,-0.75) rectangle (-0.25,0.25);
\end{tikzpicture} &

\begin{tikzpicture}[inner sep=0]
\node at (0,0){\includegraphics[width=0.18\textwidth]{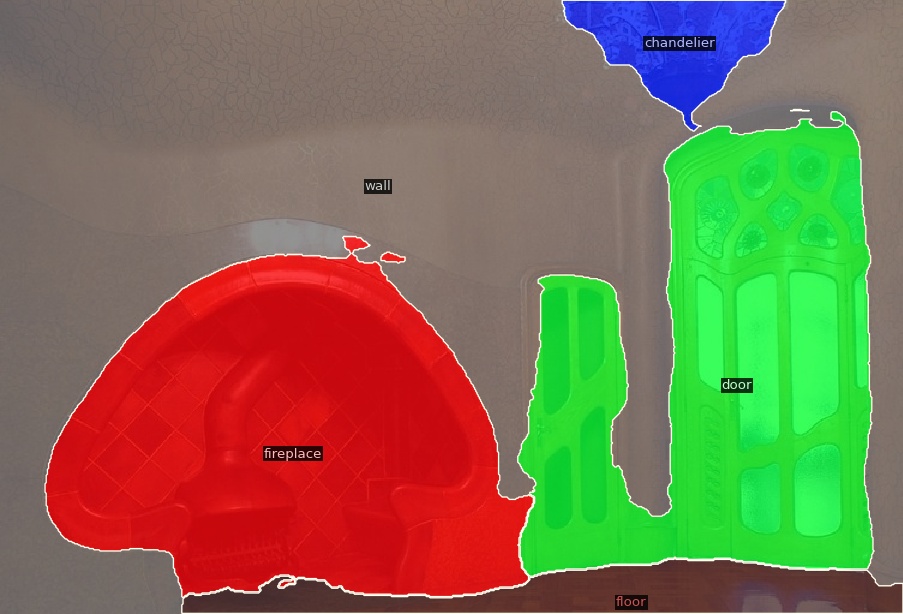}};
\draw[blue, thick, densely dashed] (-0.9,-0.75) rectangle (-0.25,0.25);
\end{tikzpicture} &

\begin{tikzpicture}[inner sep=0]
\node at (0,0){\includegraphics[width=0.18\textwidth]{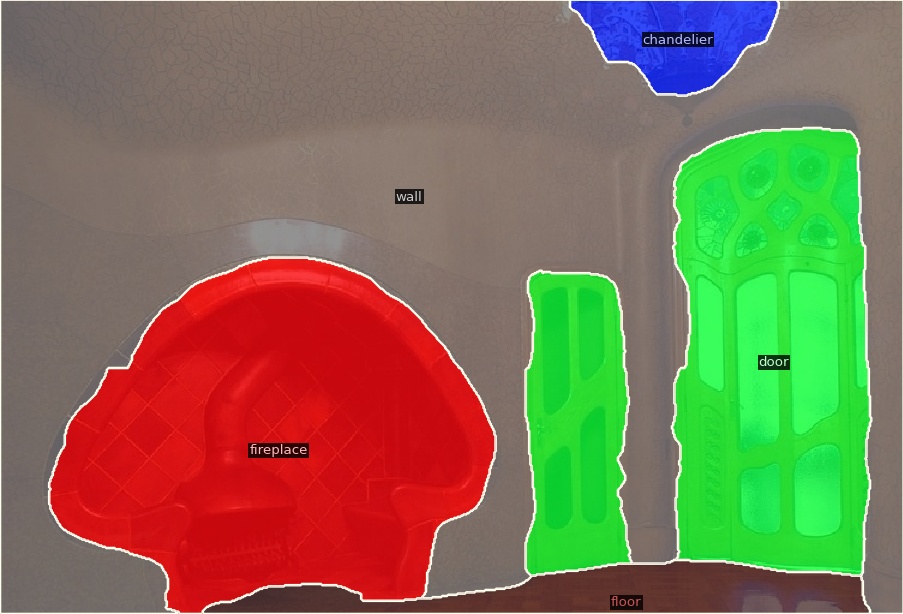}};
\draw[blue, thick, densely dashed] (-0.9,-0.75) rectangle (-0.25,0.25);
\end{tikzpicture} &

\begin{tikzpicture}[inner sep=0]
\node at (0,0){\includegraphics[width=0.18\textwidth]{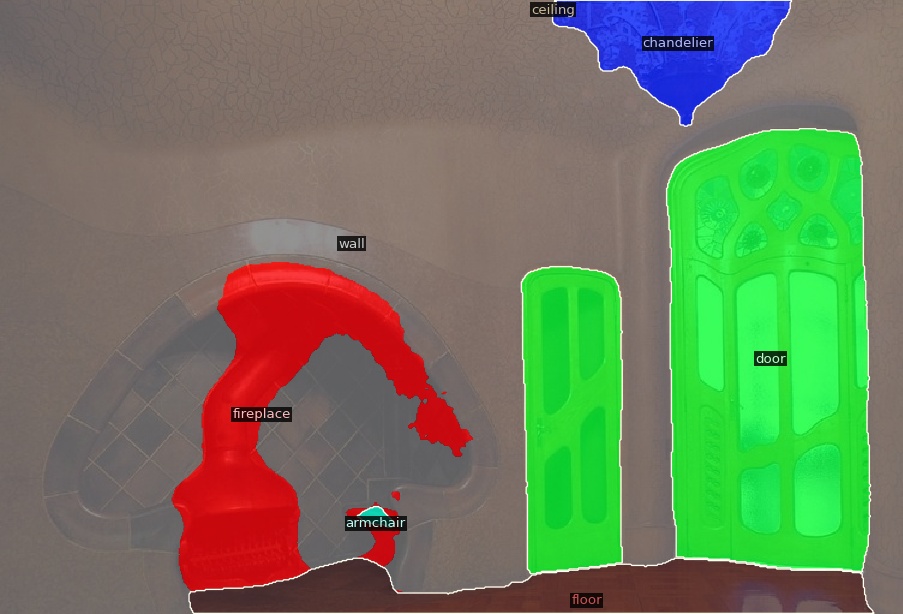}};
\draw[blue, thick, densely dashed] (-0.9,-0.75) rectangle (-0.25,0.25);
\end{tikzpicture} &

\begin{tikzpicture}[inner sep=0]
\node at (0,0){\includegraphics[width=0.18\textwidth]{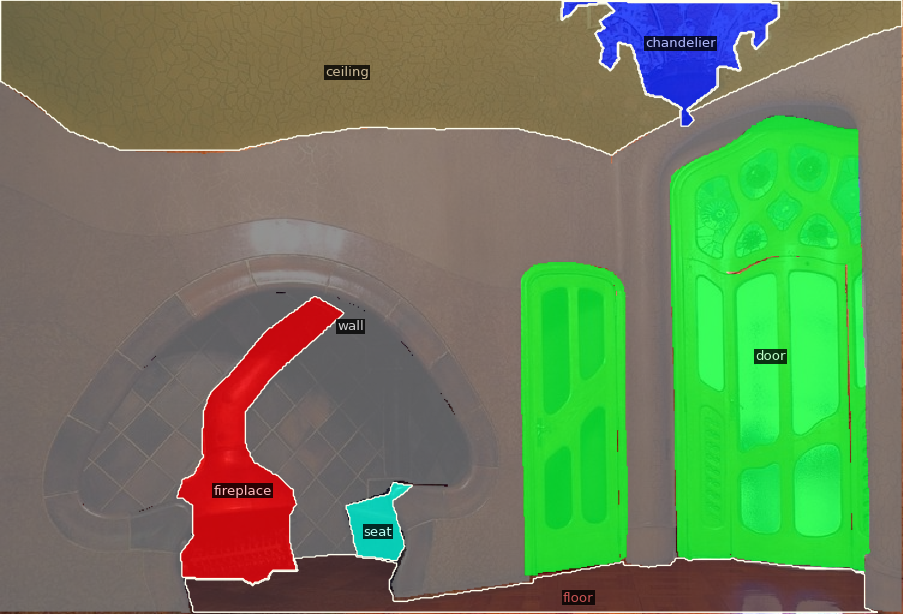}};
\draw[blue, thick, densely dashed] (-0.9,-0.75) rectangle (-0.25,0.25);
\end{tikzpicture} \\

\multicolumn{1}{c}{\small Input Image} &
\multicolumn{1}{c}{\small SED} &
\multicolumn{1}{c}{\small CAT-Seg} &
\multicolumn{1}{c}{\small Ours} &
\multicolumn{1}{c}{\small Ground Truth} \\[-2pt]

\end{tabular}

\caption{\textbf{Qualitative comparison with state-of-the-art.} Highlighted regions in bounding boxes illustrate cases where our model achieves more accurate segmentation results.}
\label{qual_results}
\end{figure*}
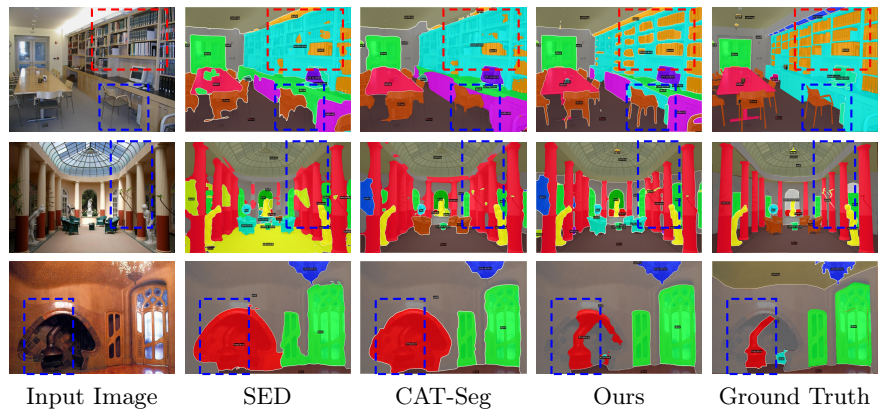

Fig.~\ref{qual_results} provides qualitative comparisons with state-of-the-art approaches. Our method produces more accurate segmentations for both seen categories (e.g., ``chair'' and ``book'') and unseen categories (e.g., ``column'' and ``fireplace''), further demonstrating improved generalization across diverse classes. Additional qualitative results are provided in the supplementary material.

\subsection{Ablation Study}
\label{subsec:ablations}

\textbf{Ablation Study of Design Components.}
Table~\ref{tab:config_ablation_updated} presents a systematic evaluation of individual design components. We first report two reference baselines using \texttt{dinov3.txt} — a zero-shot variant (ZS) without any task-specific training, and a fine-tuned variant (FT) trained with Binary Cross-Entropy (BCE) loss — both following the inference protocol of~\cite{catseg} and relying solely on local text embeddings for correlation map computation. While the fine-tuned variant shows notable gains over its zero-shot counterpart, there remains significant room for improvement, which we address through the design choices explored in the following configurations.

Building on the fine-tuned baseline, Config~I strengthens the setup by incorporating established practices in OVSS \cite{catseg, dutta2025aeroseg}, namely SAM-guided correlation feature refinement and an upsampling decoder. This modification consistently improves performance across all datasets, establishing a stronger reference for subsequent ablations.

Introducing a text embedding ensemble that combines global and local text representations (Config~II) yields consistent gains on most benchmarks, particularly on A-847, PC-459, and PC-59, indicating improved class-level semantic alignment. Adding early VLM feature refinement (Config~III) further improves performance on ADE benchmarks (A-847 and A-150), highlighting the benefit of refining dense visual features at earlier stages. Replacing BCE with a weighted focal-dice objective (Config~IV) provides additional improvements, especially on A-847 and PC-459.

Finally, incorporating the proposed LGA-based inference strategy (Config~V) improves results on ADE datasets and PC-459 while maintaining comparable performance on the remaining benchmarks, demonstrating the complementary effects of the proposed training and inference components.

\setlength{\tabcolsep}{4pt} 
\begin{table*}[!h]
\centering
\renewcommand{\arraystretch}{1}
\caption{\textbf{Ablation study on various design choices.}
Text Ens. = Text Embedding Ensemble; Early Ref. = Early Refinement; LGA Inf. = LGA-based Inference Strategy.}
\label{tab:config_ablation_updated}
\resizebox{\linewidth}{!}{
\begin{tabular}{c cccc ccccc}
\toprule
\textbf{Config.}
& \multicolumn{4}{c}{}
& \textbf{A-847} & \textbf{PC-459} & \textbf{A-150} & \textbf{PC-59} & \textbf{PAS-20} \\
\midrule
\rowcolor{gray!20}
\multicolumn{10}{l}{\textit{\textbf{Reference Baselines}}} \\
\texttt{dinov3.txt} ZS & & & & & 8.26  & 8.82  & 18.12 & 25.94 & 82.31 \\
\texttt{dinov3.txt} FT & & & & & 8.86  & 17.46 & 28.33 & 59.64 & 96.69 \\
\rowcolor{gray!20}
\multicolumn{10}{l}{\textit{\textbf{Ablation Configurations}}} \\
& \textit{Text Ens.} & \textit{Early Ref.} & \textit{Focal-Dice} & \textit{LGA Inf.}
& & & & & \\[-2pt]
\cmidrule(lr){2-5}
\textbf{I}
& $\times$      & $\times$      & $\times$      & $\times$
& 18.70 & 26.51 & 41.57 & 64.05 & 97.61 \\
\textbf{II}
& $\checkmark$  & $\times$      & $\times$      & $\times$
& 19.22 & 27.29 & 41.45 & 64.20 & 97.75 \\
\textbf{III}
& $\checkmark$  & $\checkmark$  & $\times$      & $\times$
& 19.44 & 27.23 & 41.84 & 64.22 & 97.64 \\
\textbf{IV}
& $\checkmark$  & $\checkmark$  & $\checkmark$  & $\times$
& 19.67 & 27.45 & 41.84 & \textbf{64.33} & \textbf{97.88} \\
\textbf{V}
& $\checkmark$  & $\checkmark$  & $\checkmark$  & $\checkmark$
& \textbf{20.09} & \textbf{27.80} & \textbf{42.19} & 64.27 & 97.86 \\
\bottomrule
\end{tabular}
}
\end{table*}

\textbf{Fine-tuning Strategy Analysis.}
Following \cite{catseg}, we study different finetuning strategies for adapting the VLM backbone.
Specifically, we evaluate: (i) \textbf{QV FT}, where only the query and value projection matrices of the transformer layers are updated; (ii) \textbf{QK FT}, where the query and key projection matrices are trained while keeping the remaining parameters frozen; and (iii) \textbf{Full FT}, which finetunes all backbone parameters end-to-end.
All variants are evaluated using our Config~I model.
As shown in Table~\ref{tab:finetune_strategy}, full finetuning consistently outperforms partial finetuning strategies across most benchmarks.

\begin{table}[!h]
\centering
\vspace{-10px}
\renewcommand{\arraystretch}{0.9}
\caption{Comparison across finetuning strategies.}
\label{tab:finetune_strategy}
\resizebox{0.7\linewidth}{!}{%
\begin{tabular}{l c c c c c}
\toprule
\textbf{FT Strategy}
& \textbf{A-847} & \textbf{PC-459} & \textbf{A-150}
& \textbf{PC-59} & \textbf{PAS-20} \\
\midrule
QV FT
& 17.72 & 26.22 & 40.62 & 62.63 & 97.27 \\

QK FT
& 17.92 & \textbf{26.61} & 40.26 & 62.95 & 97.32 \\

\textbf{Full FT (Ours)}
& \textbf{18.70} & 26.51 & \textbf{41.57} & \textbf{64.05} & \textbf{97.61} \\
\bottomrule
\end{tabular}
}
\vspace{-15px}
\end{table}

\textbf{Ablation on Local-Global Aggregation (LGA).}
We analyze the impact of Local--Global Aggregation (LGA) at inference time by selectively enabling it for the \vlm\ vision encoder and the SAM semantic prior encoder. When LGA is disabled for a module, the resized high-resolution image is directly fed into the corresponding encoder. As shown in Table~\ref{tab:ablation_lga}, enabling LGA for the \vlm\ vision encoder consistently improves performance over the global-only baseline across most benchmarks. In contrast, applying LGA to the SAM encoder does not yield additional gains compared to global-only inference, suggesting that high-resolution global guidance features are already sufficient. Therefore, we adopt the configuration with LGA enabled for \vlm\ and disabled for SAM as the final inference strategy, as it provides the best balance between accuracy and inference efficiency.

\begin{table}[!h]
\vspace{-15px}
\centering
\caption{Ablation on Local-Global Aggregation.}
\label{tab:ablation_lga}
\renewcommand{\arraystretch}{0.9}
\resizebox{0.8\linewidth}{!}{%
\begin{tabular}{cc ccccc @{\hspace{6pt}} | c}
\toprule
\multicolumn{2}{c}{\textbf{LGA}} &
\multicolumn{6}{c}{\textbf{Performance}} \\
\cmidrule(lr){1-2}\cmidrule(lr){3-8}
\textbf{\vlm} & \textbf{SAM} &
\textbf{A-847} & \textbf{PC-459} & \textbf{A-150} & \textbf{PC-59} & \textbf{PAS-20} & \textbf{Avg} \\
\midrule
$\times$ & $\times$ &
20.06 & 27.43 & 42.14 & 63.94 & 98.34 & 50.38 \\
$\times$ & $\checkmark$ &
20.02 & 27.49 & 42.03 & 64.07 & \textbf{98.37} & 50.40 \\
$\checkmark$ & $\times$ &
\textbf{20.09} & 27.80 & \textbf{42.19} & 64.27 & 97.86 & \textbf{50.44} \\
$\checkmark$ & $\checkmark$ &
19.95 & \textbf{27.83} & 42.10 & \textbf{64.32} & 97.94 & 50.43 \\
\bottomrule
\end{tabular}
}
\vspace{-15px}
\end{table}

\section{Conclusion}

In this paper, we introduced \texttt{dinov3.seg}, a dedicated OVSS framework built on top of \texttt{dinov3.txt} that explicitly optimizes vision–language representations for dense, segmentation-aware prediction. By integrating complementary global and local textual embeddings, dual-stage refinement of visual features and image–text correlations, and a high-resolution local–global inference strategy, our approach strengthens semantic alignment while preserving fine-grained spatial structure and boundary fidelity. Extensive experiments across five challenging OVSS benchmarks demonstrate consistent and state-of-the-art performance, highlighting the importance of segmentation-specific adaptation of DINO-based VLMs for robust open-vocabulary dense understanding.


%
%
\bibliographystyle{splncs04}
\bibliography{main}

\clearpage
\setcounter{section}{0}
\section{Outline of Supplementary Material}

This supplementary material offers further insights and expanded analysis to complement the main paper. The key sections are outlined below:

\begin{itemize}
    \item \textbf{(Sec. 2) Analysis of Local and Global Textual Embeddings:} Provides quantitative and qualitative analysis on the complementary nature of local and global textual embeddings.
    \item \textbf{(Sec. 3) Seen and Unseen Class Performance Evaluation:} Details how the seen-unseen split is constructed for benchmark datasets, along with corresponding results.
    \item \textbf{(Sec. 4) Additional Ablation Studies:} We present ablation analysis on: (i) the number of late refinement blocks used, (ii) different SPE choices, (iii) different VLM choices, and (iv) loss function hyperparameters, (v) qualitative analysis of LGA-based inference strategy.
    \item \textbf{(Sec. 5) Comparison with CAT-Seg equipped with \texttt{dinov3.txt}:} Compares our approach against CAT-Seg  with \texttt{dinov3.txt} VLM backbone.
    \item \textbf{(Sec. 6) Complexity Analysis:} Examines the computational complexity of the proposed method and provides comparisons with various state-of-the-art models.
    \item \textbf{(Sec. 7) Additional Qualitative Results:} Presents further qualitative examples to complement the results reported in the main paper.
\end{itemize}

\section{Analysis on Local and Global textual embeddings}

\subsection{Quantitative Analysis: Text-to-Visual Prototype Predictability}
\label{glocal_quant}

To quantitatively assess whether Global Text Embeddings and Local Text Embeddings contribute complementary information with respect to the visual space, we train a  Ridge regression model to predict the visual prototype embeddings from the text embeddings and measure the coefficient of determination ($R^2$) via 5-fold cross-validation. 
We extract visual prototype embeddings for each class by computing the class-wise average of mask-pooled pretrained DINOv3 features. We consider three configurations: Global Text Embeddings alone, Local Text Embeddings alone, and the concatenation of both. All embeddings are L2-normalized prior to regression. Results are reported in Table~\ref{tab:r2}.

\begin{table}[h]
\centering
\small
\caption{$R^2$ scores for predicting DINOv3 visual prototypes from text embeddings 
using Ridge regression.}
\label{tab:r2}
\begin{tabular}{l@{\hspace{1cm}}c}
\toprule
\textbf{Configuration} & \textbf{$R^2$} \\
\midrule
Global Text Embeddings only  & 0.0987 \\
Local Text Embeddings only   & 0.1376 \\
Both (concatenated)          & 0.1718 \\
\bottomrule
\end{tabular}
\end{table}

Both Global Text Embeddings and Local Text Embeddings individually explain a meaningful portion of the visual prototype space, achieving $R^2$ scores of 0.0987 and 0.1376 
respectively. More importantly, combining both yields an $R^2$ of 0.1718, which is substantially higher than either embedding alone. This improvement demonstrates that Global Text Embeddings and Local Text Embeddings capture complementary visual-semantic  correspondences, and that both are necessary to better account for the structure of the visual space.

\subsection{Qualitative Analysis: Nearest-Neighbor Complementarity}

We inspect the nearest neighbors of individual classes in the Global Text Embedding space, Local Text Embedding space, and the Visual prototype space extracted using DINOv3, and present two representative examples in Table~\ref{tab:nn}.

\begin{table}[h]
\vspace{-10px}
\centering
\small
\setlength{\tabcolsep}{10pt}
\renewcommand{\arraystretch}{1.3}
\caption{Top-5 nearest neighbors in each embedding space for two representative classes in the A-150 dataset.
Neighbors matching the visual space are \textbf{bolded}.}
\label{tab:nn}
\resizebox{\textwidth}{!}{
\begin{tabular}{l l p{7cm}}
\toprule
\textbf{Class} & \textbf{Embedding Space} & \textbf{Top-5 Nearest Neighbors} \\
\midrule
\multirow{3}{*}{\textit{Stairway}}
 & Visual Prototypes      & stairs, bannister, railing, step, grandstand \\
 & Global Text Embeddings & \textbf{stairs}, \textbf{bannister}, path, \textbf{step}, \textbf{railing} \\
 & Local Text Embeddings  & \textbf{stairs}, escalator, \textbf{bannister}, path, \textbf{railing} \\
\midrule
\multirow{3}{*}{\textit{Minibike}}
 & Visual Prototypes      & bicycle, car, van, truck, ashcan \\
 & Global Text Embeddings & \textbf{bicycle}, \textbf{car}, road, path, person \\
 & Local Text Embeddings  & \textbf{bicycle}, chair, \textbf{truck}, \textbf{car}, bench \\
\bottomrule
\end{tabular}
}
\vspace{-10px}
\end{table}

For \textit{stairway}, Global Text Embeddings retrieve more neighbors consistent with 
the visual space by capturing structural parts such as \textit{bannister}, \textit{railing}, 
and \textit{step}, while Local Text Embeddings introduce \textit{escalator} --- an object that shares functional purpose but differs visually. For \textit{minibike}, Local Text Embeddings recover three visually consistent object-level neighbors, whereas Global Text Embeddings retrieves scene-level elements (\textit{road}, \textit{path}) that 
reflect the typical environment of a minibike rather than its appearance. In both cases, the two embedding spaces make different errors and different correct retrievals, suggesting that Global Text Embeddings and Local Text Embeddings encode complementary semantic information that together better covers the structure of the visual embedding space.

\section{Seen and Unseen Class Performance Evaluation}

\vspace{-10px}

Since the benchmark datasets used in this work do not provide an explicit seen–unseen class split, we approximate this partition using two complementary similarity measures: (i) visual feature similarity, and (ii) textual embedding similarity.

\vspace{-10px}

\subsection{Visual Feature Similarity-Based Seen–Unseen Class Partition}
\begin{figure}
\vspace{-20px}
    \centering
    \includegraphics[width=0.75\linewidth]{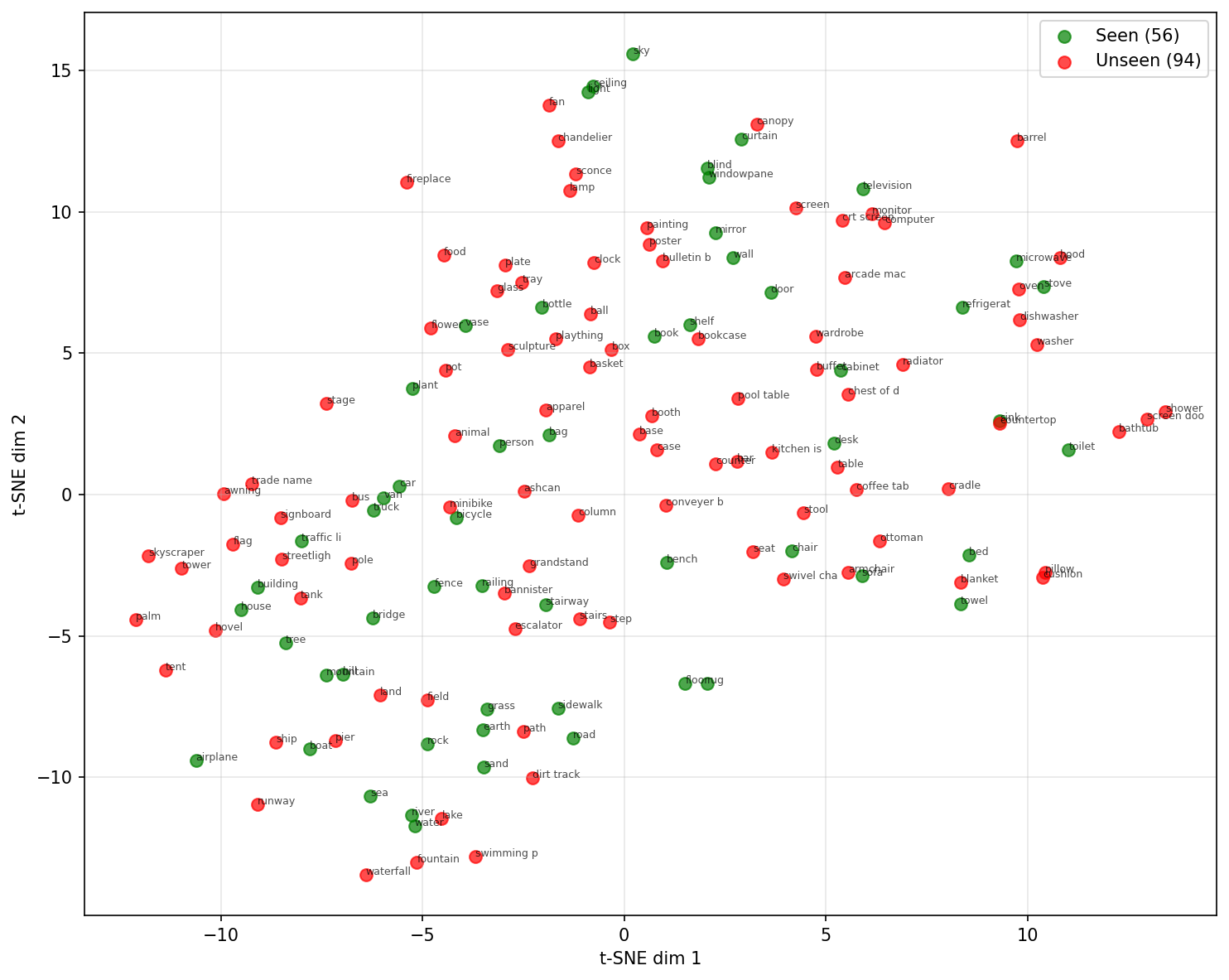}
    \caption{t-SNE visualization of visual prototypes for A-150, with seen classes (green) and unseen classes (red).}
    \label{fig:tsne_vis}
    \vspace{-10px}
\end{figure}

We extract a visual prototype for each training and test class using pretrained DINOv3, as described in Sec.~\ref{glocal_quant}. A test class is designated as \textit{seen} if its cosine similarity to at least one training-class visual prototype exceeds a threshold of 0.9; otherwise, it is considered \textit{unseen}. The resulting class counts under this partition are reported in Table~\ref{seen_unseen_split}, and the corresponding mIoU scores on these subsets are presented in Table~2 of the main paper. Figure~\ref{fig:tsne_vis} shows a t-SNE visualization of visual prototypes for A-150 dataset.

\begin{table}[!h]
\vspace{-10px}
\caption{Seen and unseen class splits for different datasets based on visual similarity. }
\centering
\begin{tabular}{l cc}
\hline
Dataset & \# Seen  & \# Unseen  \\
\hline
A-847 & 72 & 775 \\
PC-459 \footnotemark  & 96 & 249 \\
A-150 & 56 & 94 \\
PC-59 & 56 & 3 \\
PAS-20 & 20 & 0 \\
\hline
\end{tabular}
\vspace{-10px}
\label{seen_unseen_split}
\end{table}

\footnotetext{PC-459 has 114 absent classes which are not considered in our analysis.}

\subsection{Textual Embeddings Similarity-based Seen-Unseen Class Partition.}

\begin{figure}[!h]
    \centering
    \includegraphics[width=0.75\linewidth]{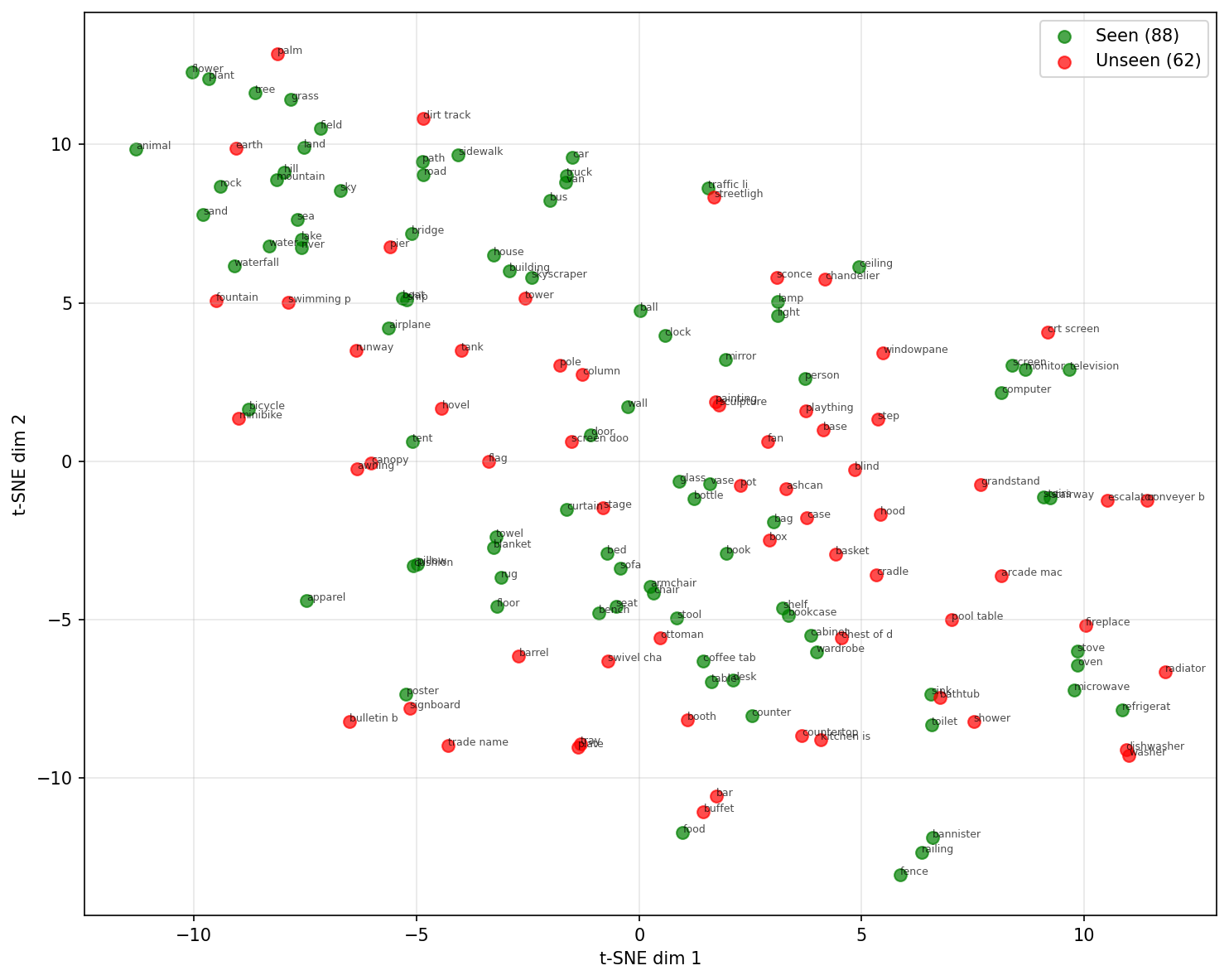}
    \caption{t-SNE visualization of textual embeddings for A-150, with seen classes (green) and unseen classes (red).}
    \label{fig:tsne_text}
\end{figure}

To construct the seen–unseen partition based on textual similarity, we extract global and local text embeddings from pretrained \vlm\ text encoder and concatenate them to form a class-level representation for each training and test class. A test class is designated as \textit{seen} if its cosine similarity to at least one training-class textual representation exceeds a threshold of 0.9; otherwise, it is considered \textit{unseen}. The resulting class counts under this partition are reported in Table~\ref{tab:seen_unseen_split_text}. Figure~\ref{fig:tsne_text} presents a t-SNE visualization of textual embeddings on the A-150 dataset.

\begin{table}[!h]
\centering
\vspace{-10px}
\caption{Seen and unseen class splits for different datasets  based on textual similarity.}
\begin{tabular}{l cc}
\hline
Dataset & \# Seen  & \# Unseen  \\
\hline
A-847 & 196 & 651 \\
PC-459 & 119 & 226 \\
A-150 & 88 & 62 \\
PC-59 & 56 & 3 \\
PAS-20 & 20 & 0 \\
\hline
\end{tabular}
\vspace{-10px}
\label{tab:seen_unseen_split_text}
\end{table}

We present seen and unseen mIoU scores based on this split in Table~\ref{tab:miou_text_split}. Our method achieves the highest mIoU on both seen and unseen classes across A-847, PC-459, and A-150, with gains most pronounced on unseen categories — up to +5.90 mIoU over CAT-Seg on A-150 — demonstrating strong generalization to novel classes. On PC-59, our method improves seen class performance but shows a marginal drop of 0.28 mIoU on unseen classes, possibly owing to the very few unseen classes (only 3) in that split.

\begin{table}[ht]
\centering
\begin{tabular}{l cc cc cc cc}
\hline
 & \multicolumn{2}{c}{\textbf{A-847}} 
 & \multicolumn{2}{c}{\textbf{PC-459}} 
 & \multicolumn{2}{c}{\textbf{A-150}}
 & \multicolumn{2}{c}{\textbf{PC-59}} \\
Method 
 & Seen & Unseen 
 & Seen & Unseen 
 & Seen & Unseen
 & Seen & Unseen \\
\hline
SED \cite{xie2024sed}
 & 18.98 & 12.35
 & 35.27 & 15.99
 & 40.30 & 27.83
 & 61.43 & \textbf{45.97} \\
MAFT+ \cite{maftp}
 & 21.59 & 13.66
 & 34.80 & 13.98
 & 41.93 & 28.39
 & 61.26 & 26.52 \\
CAT-Seg \cite{catseg}
 & 22.60 & 13.90
 & 38.99 & 15.73
 & 42.22 & 31.51
 & 61.43 & \textbf{45.97} \\
\texttt{dinov3.seg} (Ours)
 & \textbf{25.57} & \textbf{18.49}
 & \textbf{40.63} & \textbf{21.04}
 & \textbf{45.55} & \textbf{37.42}
 & \textbf{65.26} & 45.69 \\
\hline
$\Delta$
 & +2.97 & +4.58
 & +1.64 & +5.31
 & +3.33 & +5.90
 & +3.83 & -0.28 \\
\hline
\end{tabular}
\caption{Seen and unseen class (based on textual similarity) mIoU comparison across datasets. $\Delta$ indicates absolute improvement over CAT-Seg.}
\label{tab:miou_text_split}
\end{table}

\section{Additional ablation studies}

\subsection{Ablation on number of late refinement blocks} 

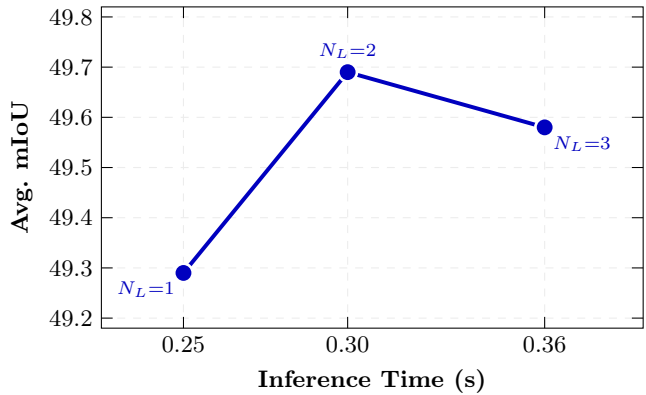
\begin{figure}[!htp]
\centering
\begin{tikzpicture}
\begin{axis}[
    width=0.72\linewidth,
    height=0.48\linewidth,
    xlabel={Inference Time (s)},
    ylabel={Avg. mIoU},
    xmin=0.225, xmax=0.39,
    ymin=49.18, ymax=49.82,
    axis background/.style={fill=white},
    grid=major,
    grid style={dashed, gray!15},
    xtick={0.25, 0.30, 0.36},
    xticklabels={0.25, 0.30, 0.36},
    ytick distance=0.1,
    scaled ticks=false,
    tick label style={font=\small},
    label style={font=\small, font=\bfseries},
    tick style={black, thin},
    axis line style={black, thin},
    enlargelimits=false,
    clip=false,
]

\addplot[
    color=blue!75!black,
    line width=1.5pt,
    mark=*,
    mark size=3.5pt,
    mark options={fill=blue!75!black, draw=white, line width=1pt}
]
coordinates {
    (0.25,49.29)
    (0.30,49.69)
    (0.36,49.58)
};

\node[below left, font=\scriptsize\bfseries, text=blue!75!black]
    at (axis cs:0.25,49.29) {$N_L{=}1$};
\node[above, font=\scriptsize\bfseries, text=blue!75!black, yshift=2pt]
    at (axis cs:0.30,49.69) {$N_L{=}2$};
\node[below right, font=\scriptsize\bfseries, text=blue!75!black]
    at (axis cs:0.36,49.58) {$N_L{=}3$};

\end{axis}
\end{tikzpicture}
\caption{Accuracy--runtime trade-off for varying numbers of late refinement
blocks~$N_L$. $N_L{=}2$ achieves the best average mIoU (49.69) with a
moderate inference cost of 0.30\,s.}
\label{fig:miou_runtime}
\end{figure}

Table~\ref{num_refine} reports the effect of the number of late refinement blocks ($N_L$) in our framework, where each late refinement block consists of a spatial refinement block and a class refinement block. Using a single refinement block ($N_L$=1) leads to noticeably weaker performance on A-150 and A-847, while results on PAS-20 remain high, indicating limited sensitivity on simpler datasets. Increasing $N_L$ to 2 yields consistent improvements across most datasets, giving the best overall performance. Further increasing $N_L$ to 3 results in a slight drop on most datasets, suggesting over-refinement. As shown in Fig.~\ref{fig:miou_runtime}, $N_L$=2 strikes the best balance between inference time and segmentation performance, and is therefore chosen as the default setting throughout the paper.

\begin{table}[!htp]
\centering
\caption{Effect of number of late refinement blocks.}
\setlength{\tabcolsep}{4pt}
\begin{tabular}{c|ccccc|c}
\toprule
\textbf{$N_L$} & \textbf{A-847} & \textbf{PC-459} & \textbf{A-150} & \textbf{PC-59} & \textbf{PAS-20} & \textbf{Avg.} \\
\midrule
1 & 18.09 & \textbf{26.52} & 40.19 & 63.85 & \textbf{97.81} & 49.29 \\
\rowcolor{gray!15} 2 & \textbf{18.70} & 26.51 & \textbf{41.57} & 64.05 & 97.61 & \textbf{49.69} \\
3 & 18.43 & 26.28 & 41.34 & \textbf{64.09} & 97.75 & 49.58 \\
\bottomrule
\end{tabular}
\label{num_refine}
\end{table}

\subsection{Exploration with different SPEs}

Table~\ref{tab:spe_comparison} reports the effect of different Semantic Prior Encoder (SPE) choices on segmentation performance, with all experiments conducted using our Config~I model. 
In the \emph{none} setting, DINOv3 features from the \vlm\ backbone are used directly as semantic prior features. Incorporating a dedicated SPE consistently improves over this baseline, confirming the benefit of explicit semantic priors. Among all variants, SAM ViT-L achieves the best average mIoU of 49.69 with top scores on three out of five benchmarks, and is therefore set as our default SPE throughout the paper. While SAM3 PE-L+ and SAM-2.1 Hiera-B+ are competitive on some datasets, neither matches the overall consistency of SAM ViT-L.

\begin{table}[!h]
\centering
\caption{Comparison of different Semantic Prior Encoders (SPE).}

\setlength{\tabcolsep}{2pt}
\begin{tabular}{l|ccccc|c}
\hline
\textbf{SPE} & \textbf{A-847} & \textbf{PC-459} & \textbf{A-150} & \textbf{PC-59} & \textbf{PAS-20} & \textbf{Avg.} \\
\hline
\emph{none}        & 18.23 & 26.15 & 40.30 & 63.04 & 97.76 & 49.10 \\
SAM-2.1 Hiera-B+   & 18.33 & \textbf{26.86} & 41.04 & 63.77 & 97.70 & 49.54 \\
SAM-2.1 Hiera-L    & 18.39 & 25.89 & 40.55 & 62.84 & 97.60 & 49.05 \\
SAM3 PE-L+         & 18.51 & 26.60 & 41.16 & 63.72 & \textbf{97.84} & 49.57 \\
\rowcolor{gray!15} SAM ViT-L 
                   & \textbf{18.70} & 26.51 & \textbf{41.57} & \textbf{64.05} & 97.61 & \textbf{49.69} \\
\hline
\end{tabular}
\label{tab:spe_comparison}
\end{table}

\subsection{Exploration with different VLMs}
Apart from \vlm, we have experimented with two additional VLM backbones: CLIP (ViT-L) \cite{clip} and \texttt{dinov2.txt} \cite{jose2025dinotxt}. Since CLIP only offers \emph{global} text embeddings, we only utilize \emph{global} embeddings instead of using text ensemble. QV finetuning is used for CLIP since it produces optimal results as shown in \cite{catseg}. Table-\ref{tab:vlm_comp} compares performance across different VLM backbones. \vlm\ consistently achieves the best performance on most datasets, with notable gains on A-847, PC-459, and A-150. \texttt{dinov2.txt} remains competitive on PC59 and PAS-20, while CLIP lags behind across all benchmarks. Notably, our method with CLIP (ViT-L) backbone outperforms multiple state-of-the-art methods \cite{catseg,peng2025hyperbolic,peng2025peft} with same VLM backbone.

\begin{table}[!h]
\centering
\setlength{\tabcolsep}{4pt}
\caption{Ablation on VLM backbone choice and comparison with state-of-the-art CLIP (ViT-L) based OVSS methods.}
\begin{tabular}{l|ccccc|c}
\hline
\textbf{Method} & \textbf{A-847} & \textbf{PC-459} & \textbf{A-150} & \textbf{PC-59} & \textbf{PAS-20} & \textbf{Avg.} \\
\hline

\multicolumn{7}{c}{\textit{Ours with Different VLM Backbones}} \\
\hline
Ours (CLIP ViT-L)              & 17.82 & 26.55 & 38.48 & 64.08 & 97.24 & 48.83 \\
Ours (\texttt{dinov2.txt})     & 17.25 & 26.07 & 39.61 & \textbf{64.46} & 97.74 & 49.03 \\
Ours (\vlm)                    & \textbf{20.09} & \textbf{27.80} & \textbf{42.19} & 64.27 & \textbf{97.86} & \textbf{50.44} \\
\hline\hline
\multicolumn{7}{c}{\textit{Existing CLIP (ViT-L) based Methods}} \\
\hline
CAT-Seg~\cite{catseg}          & 16.0  & 23.8  & 37.9  & 63.3  & 97.0  & 47.60 \\
H-CLIP~\cite{peng2025peft}     & 16.5  & 24.2  & 38.4  & 64.1  & 97.7  & 48.18 \\
HyperCLIP~\cite{peng2025hyperbolic} & 16.3 & 24.1 & 38.2 & 64.2 & 98.3 & 48.22 \\
\hline
\end{tabular}

\label{tab:vlm_comp}
\end{table}

\subsection{Ablation on loss function hyperparameters} We evaluate different combinations of $\lambda$ and $\gamma$, as summarized in Table~\ref{tab:loss_hparam}. Overall, $\lambda = 0.05$ and $\gamma = 2$ achieve the best average performance and consistently score best or near-best across most datasets compared to other settings. Notably, variations in average performance across different hyperparameter choices are relatively small.

\begin{table}[!htp]
\centering
\caption{Loss function hyperparameter analysis.}
\setlength{\tabcolsep}{4pt}
\begin{tabular}{cc|ccccc|c}
\hline
$\lambda$ & $\gamma$ & \textbf{A-847} & \textbf{PC-459} & \textbf{A-150} & \textbf{PC-59} & \textbf{PAS-20} & \textbf{Avg.} \\
\hline
0.01 & 2 & 19.65 & \textbf{28.51} & 41.82 & 64.26 & 97.85 & 50.42 \\
0.1 & 2 &\textbf{20.15} & 27.84 & 42.10 & 64.08 & 97.86 & 50.40 \\
\rowcolor{gray!15} 0.05 & 2 & 20.09 & 27.80 & \textbf{42.19} & 64.27 & 97.86 & \textbf{50.44} \\
0.05 & 1 & 19.89 & 28.03 & 41.89 & \textbf{64.42} & \textbf{97.91} & 50.43 \\
0.05 & 3 & 19.93 & 27.96 & 42.10 & 64.06 & 97.86 & 50.38 \\
\hline
\end{tabular}
\label{tab:loss_hparam}
\end{table}

\subsection{Qualitative analysis of Proposed LGA-based inference strategy.}

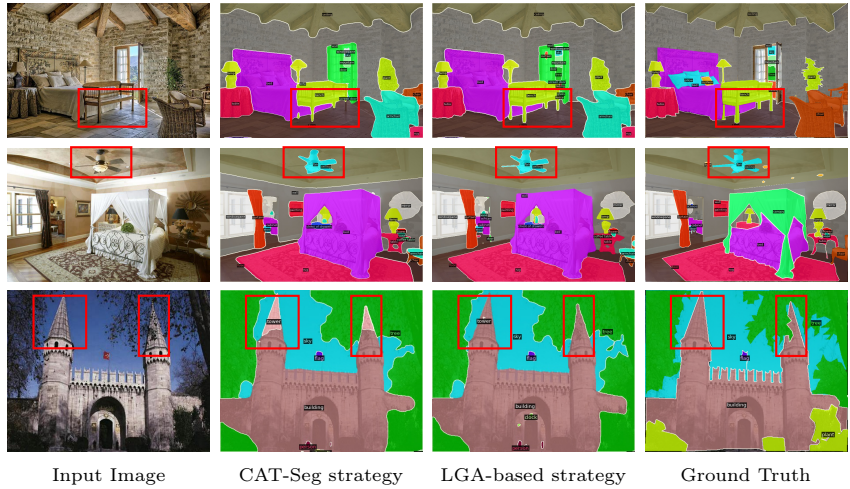
\begin{figure*}[!h]
\centering
\setlength{\tabcolsep}{2pt}
\renewcommand{\arraystretch}{1.0}
\begin{tabular}{cccc}

\begin{tikzpicture}[inner sep=0]
\node at (0,0){\includegraphics[width=0.22\textwidth]{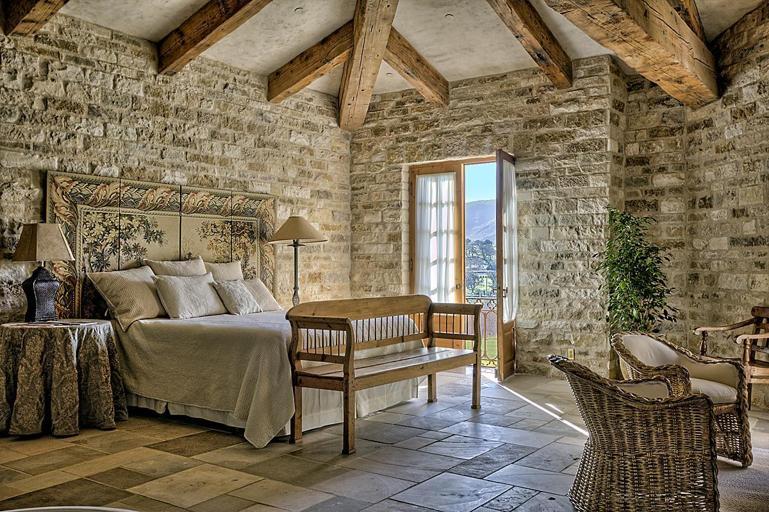}};
\draw[red, thick] (-0.4,-0.75) rectangle (0.5,-0.25);
\end{tikzpicture} &
\begin{tikzpicture}[inner sep=0]
\node at (0,0){\includegraphics[width=0.22\textwidth]{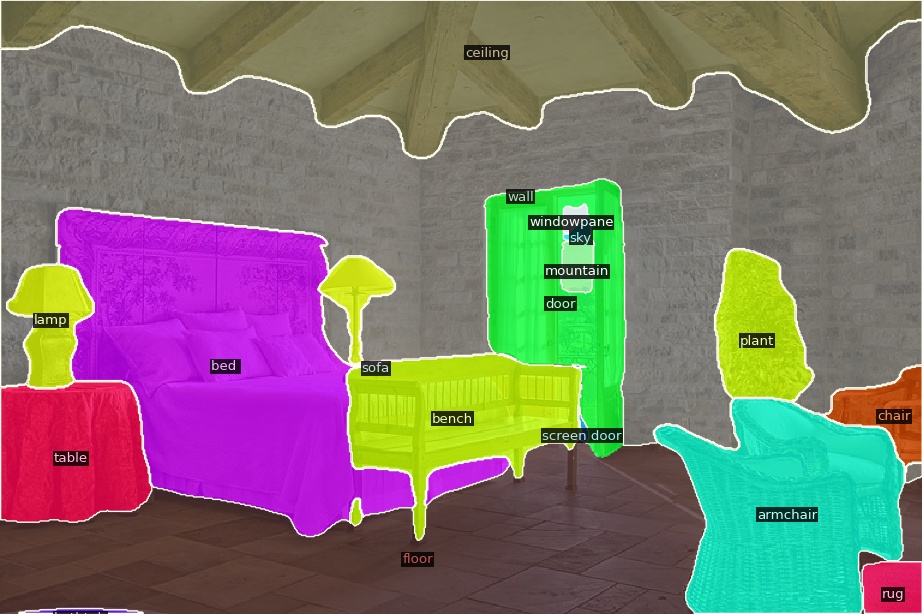}};
\draw[red, thick] (-0.4,-0.75) rectangle (0.5,-0.25);
\end{tikzpicture} &
\begin{tikzpicture}[inner sep=0]
\node at (0,0){\includegraphics[width=0.22\textwidth]{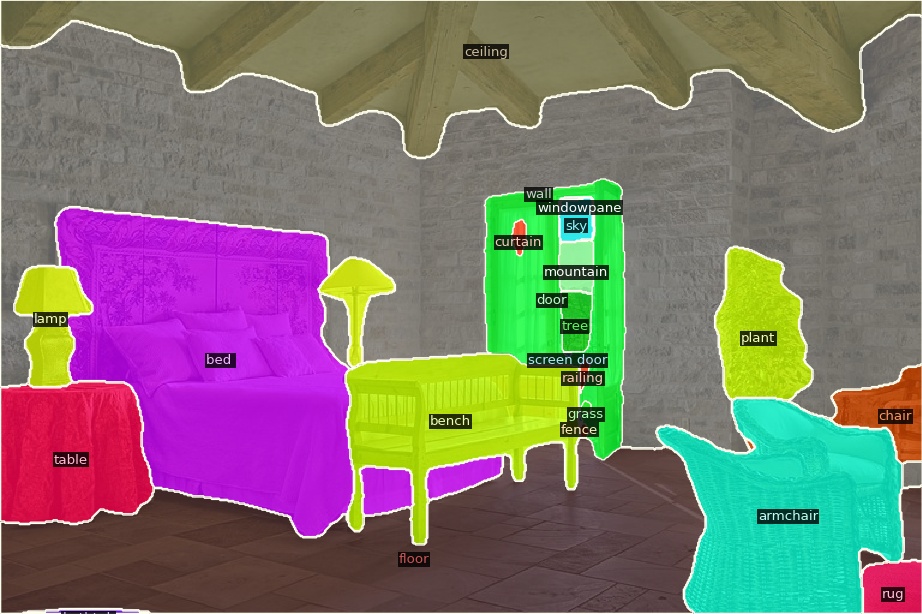}};
\draw[red, thick] (-0.4,-0.75) rectangle (0.5,-0.25);
\end{tikzpicture} &
\begin{tikzpicture}[inner sep=0]
\node at (0,0){\includegraphics[width=0.22\textwidth]{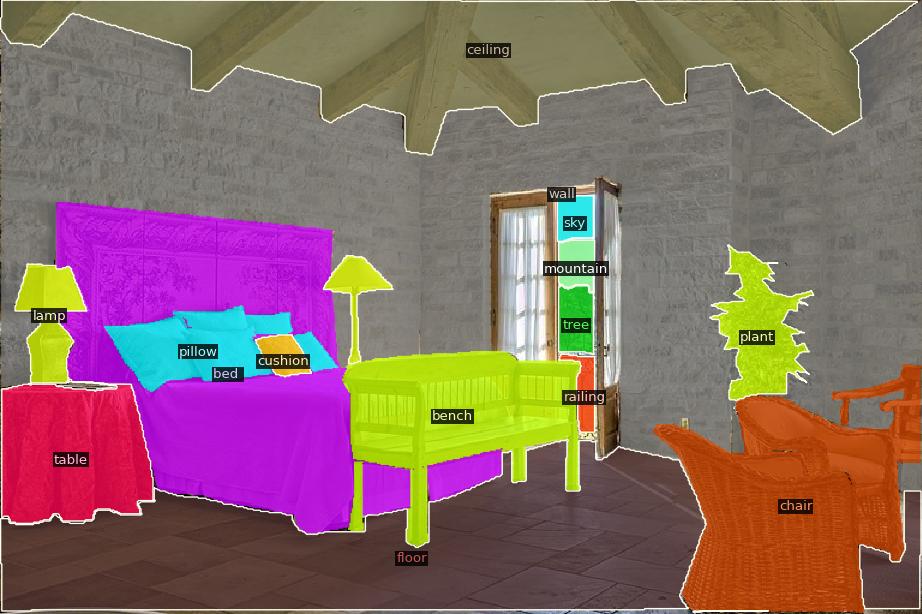}};
\draw[red, thick] (-0.4,-0.75) rectangle (0.5,-0.25);
\end{tikzpicture} \\

\begin{tikzpicture}[inner sep=0]
\node at (0,0){\includegraphics[width=0.22\textwidth]{infstr/ADE_val_00000139/image_ADE_val_00000139.jpg}};
\draw[red, thick] (-0.5,0.5) rectangle (0.3,0.9);
\end{tikzpicture} &
\begin{tikzpicture}[inner sep=0]
\node at (0,0){\includegraphics[width=0.22\textwidth]{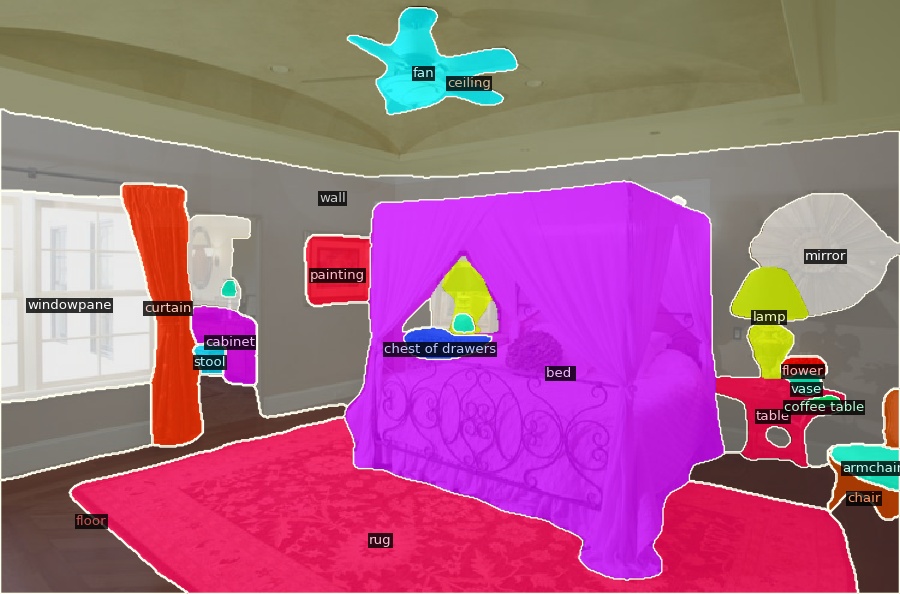}};
\draw[red, thick] (-0.5,0.5) rectangle (0.3,0.9);
\end{tikzpicture} &
\begin{tikzpicture}[inner sep=0]
\node at (0,0){\includegraphics[width=0.22\textwidth]{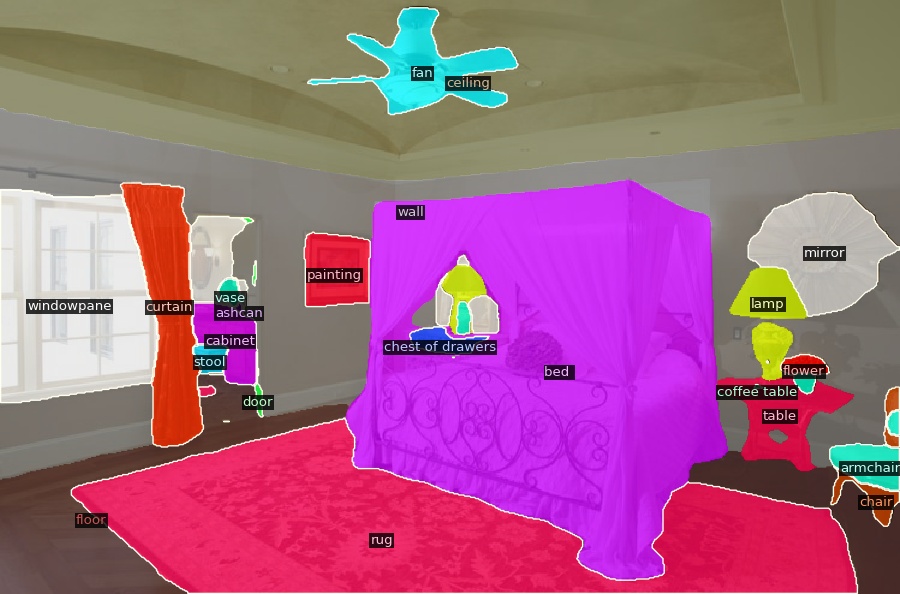}};
\draw[red, thick] (-0.5,0.5) rectangle (0.3,0.9);
\end{tikzpicture} &
\begin{tikzpicture}[inner sep=0]
\node at (0,0){\includegraphics[width=0.22\textwidth]{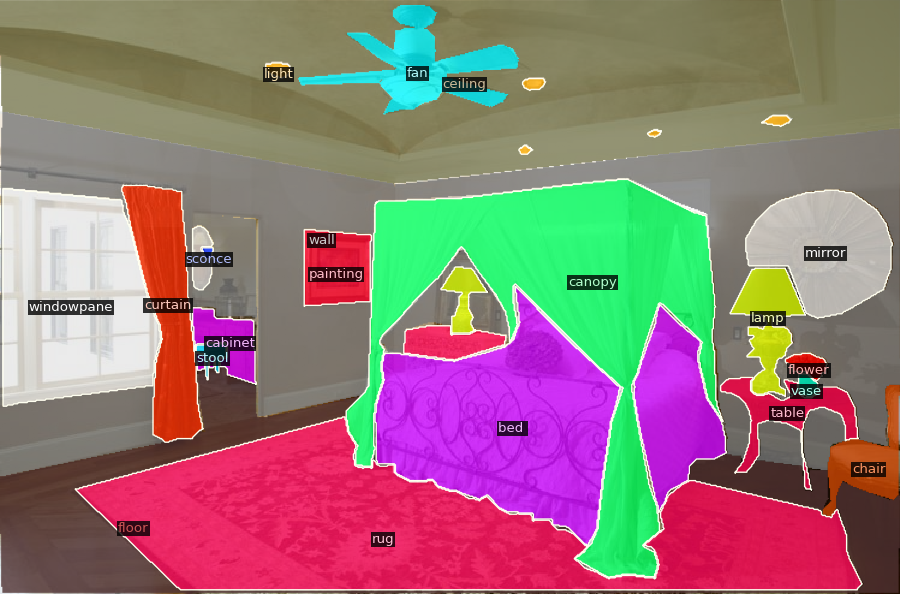}};
\draw[red, thick] (-0.5,0.5) rectangle (0.3,0.9);
\end{tikzpicture} \\

\begin{tikzpicture}[inner sep=0]
\node at (0,0){\includegraphics[width=0.22\textwidth]{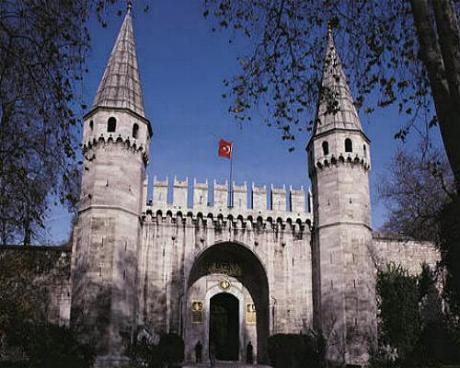}};
\draw[red, thick] (-1.0,0.3) rectangle (-0.3,1.0);
\draw[red, thick] (0.4,0.2) rectangle (0.8,1.0);
\end{tikzpicture} &
\begin{tikzpicture}[inner sep=0]
\node at (0,0){\includegraphics[width=0.22\textwidth]{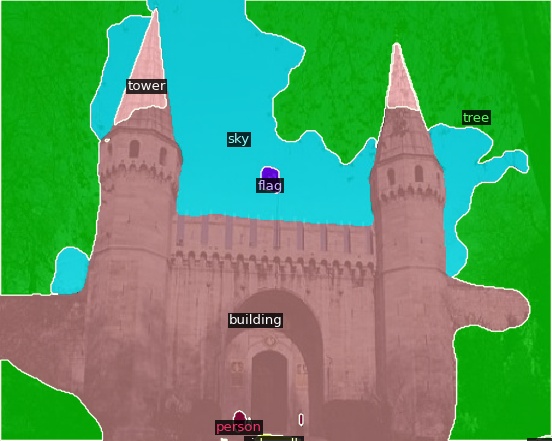}};
\draw[red, thick] (-1.0,0.3) rectangle (-0.3,1.0);
\draw[red, thick] (0.4,0.2) rectangle (0.8,1.0);
\end{tikzpicture} &
\begin{tikzpicture}[inner sep=0]
\node at (0,0){\includegraphics[width=0.22\textwidth]{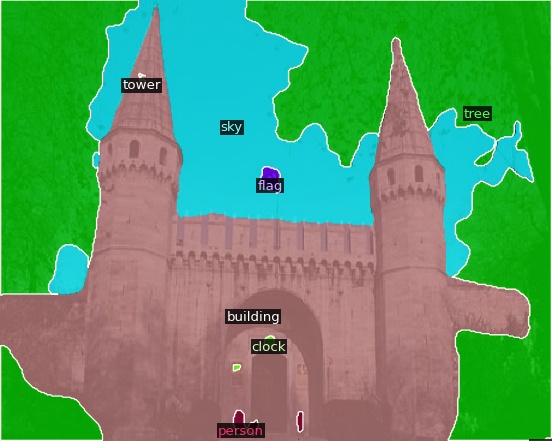}};
\draw[red, thick] (-1.0,0.3) rectangle (-0.3,1.0);
\draw[red, thick] (0.4,0.2) rectangle (0.8,1.0);
\end{tikzpicture} &
\begin{tikzpicture}[inner sep=0]
\node at (0,0){\includegraphics[width=0.22\textwidth]{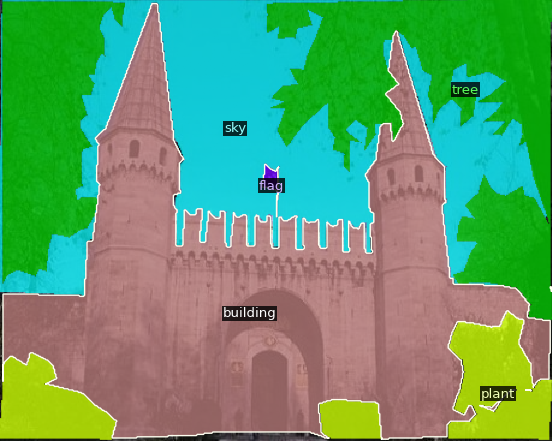}};
\draw[red, thick] (-1.0,0.3) rectangle (-0.3,1.0);
\draw[red, thick] (0.4,0.2) rectangle (0.8,1.0);
\end{tikzpicture} \\

\multicolumn{1}{c}{\scriptsize Input Image} &
\multicolumn{1}{c}{\scriptsize CAT-Seg strategy} &
\multicolumn{1}{c}{\scriptsize LGA-based strategy} &
\multicolumn{1}{c}{\scriptsize Ground Truth} \\[-2pt]

\end{tabular}
\caption{\textbf{Qualitative results on Inference Strategy. } Highlighted regions in boxes show effectiveness of proposed LGA-based strategy over CAT-Seg inference strategy.}
\label{supple_qual_infstr}
\end{figure*}

Figure~\ref{supple_qual_infstr} presents a qualitative comparison between the CAT-Seg inference strategy and our proposed LGA-based inference strategy applied to our model. Unlike CAT-Seg, our approach extracts features from both the full image and its sub-images, which are then aggregated before being passed to subsequent modules. This allows our model to capture finer-grained visual details, resulting in sharper segmentation boundaries and more accurate classification of thin and complex object structures, as illustrated in the figure.

\section{Comparison with CAT-Seg equipped with \texttt{dinov3.txt}} 

To demonstrate that the performance gains of our framework are not simply attributable to the stronger \vlm\  backbone, we retrain CAT-Seg \cite{catseg} using the same \vlm\ backbone and report results in Table~\ref{tab:catdino}. As shown, \texttt{dinov3.seg} consistently outperforms this baseline across all benchmarks, with more pronounced improvements on the larger-vocabulary datasets A-847, PC-459, and A-150. This confirms that the gains stem from our proposed segmentation framework rather than the backbone alone.

\begin{table}[!h]
\centering
\caption{Comparison w.r.t CAT-Seg with \vlm.}
\setlength{\tabcolsep}{4pt}
\begin{tabular}{lccccc|c}
\hline
\textbf{Method} & \textbf{A-847} & \textbf{PC-459} & \textbf{A-150} & \textbf{PC-59} & \textbf{PAS-20} & \textbf{Avg.} \\
\hline
CAT-Seg w/ \texttt{dinov3.txt}
 & 18.93 & 26.24 & 41.17 & 63.45 & 97.79 & 49.51 \\
\texttt{dinov3.seg}
 & \textbf{20.09} & \textbf{27.80} & \textbf{42.19} & \textbf{64.27} & \textbf{97.86} & \textbf{50.44} \\
\hline
\end{tabular}

\label{tab:catdino}
\end{table}

\clearpage

\section{Complexity Analysis}

Table~\ref{tab:efficiency} compares the computational complexity of \texttt{dinov3.seg} against existing OVSS methods. \texttt{dinov3.seg} carries a larger parameter count owing to \texttt{dinov3.txt} vision-language backbone, which accounts for 866.6M of the 1,178.2M total parameters, with the remaining 311.6M attributed to the Semantic Prior Encoder and auxiliary components. Despite this, the inference time of 0.37s remains well below that of OVSeg and SCAN (1.31s and 1.09s respectively). Notably, \texttt{dinov3.seg} achieves significantly lower GFLOPs (4,500.4) compared to both OVSeg (9,810.1) and SCAN (13,502.9), demonstrating that the increased parameter count does not translate to a proportional increase in computational cost at inference. While \texttt{dinov3.seg} is moderately heavier than CAT-Seg in terms of parameters and inference time, this overhead is modest relative to the consistent performance gains observed across all benchmarks. In future work, knowledge distillation from the Semantic Prior Encoder could be explored as a promising direction to reduce the overall computational cost of the framework.

\begin{table}[!h]
\centering
\caption{Model complexity comparison. Inference time and GFLOPs are measured on an A100 GPU at $640\times640$ resolution. GFLOPs are computed using the \texttt{fvcore} library.}
\setlength{\tabcolsep}{6pt}
\resizebox{0.8\textwidth}{!}{
\begin{tabular}{l ccc cc}
\toprule
\multirow{2}{*}{Model} 
  & \multicolumn{3}{c}{Parameters (M)} 
  & \multirow{2}{*}{Inf.\ Time (s)} 
  & \multirow{2}{*}{Inf. GFLOPs} \\
\cmidrule(lr){2-4}
 & Total & VLM & Other & & \\
\midrule
OVSeg~\cite{ovseg}         & 532.6   & 427.9 & 104.7 & 1.31 & 9{,}810.1  \\
SCAN~\cite{liu2024scan}    & 890.3   & 427.6 & 462.7 & 1.09 & 13{,}502.9 \\
CAT-Seg~\cite{catseg}      & 433.7   & 427.9 &   5.8 & 0.22 & 1{,}963.5  \\
\midrule
\texttt{dinov3.seg}        & 1{,}178.2 & 866.6 & 311.6 & 0.37 & 4{,}500.4 \\
\bottomrule
\end{tabular}
}
\label{tab:efficiency}
\end{table}

\clearpage

\section{Additional Qualitative Results}
\begin{figure*}[!h]
\centering
\vspace{-20px}
\setlength{\tabcolsep}{2pt}
\renewcommand{\arraystretch}{1.0}
\begin{tabular}{ccc}

\includegraphics[width=0.27\textwidth]{supple_qual/a847/ADE_val_00000998/image_ADE_val_00000998.jpg} &
\includegraphics[width=0.27\textwidth]{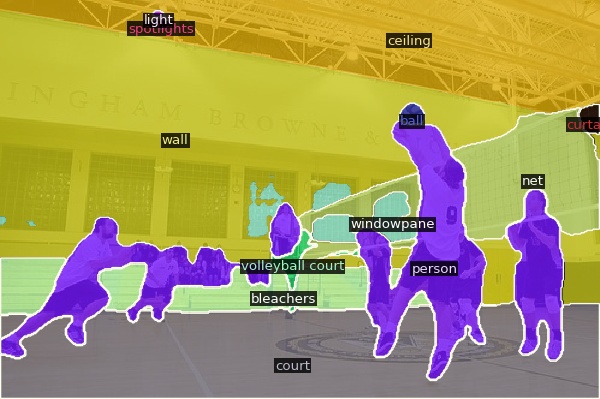} &
\includegraphics[width=0.27\textwidth]{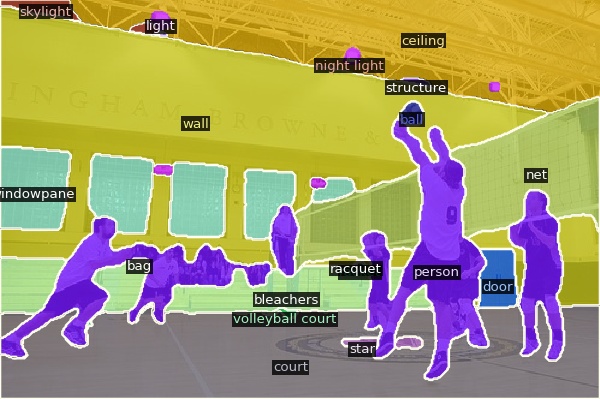} \\

\includegraphics[width=0.27\textwidth]{supple_qual/a847/ADE_val_00000314/image_ADE_val_00000314.jpg} &
\includegraphics[width=0.27\textwidth]{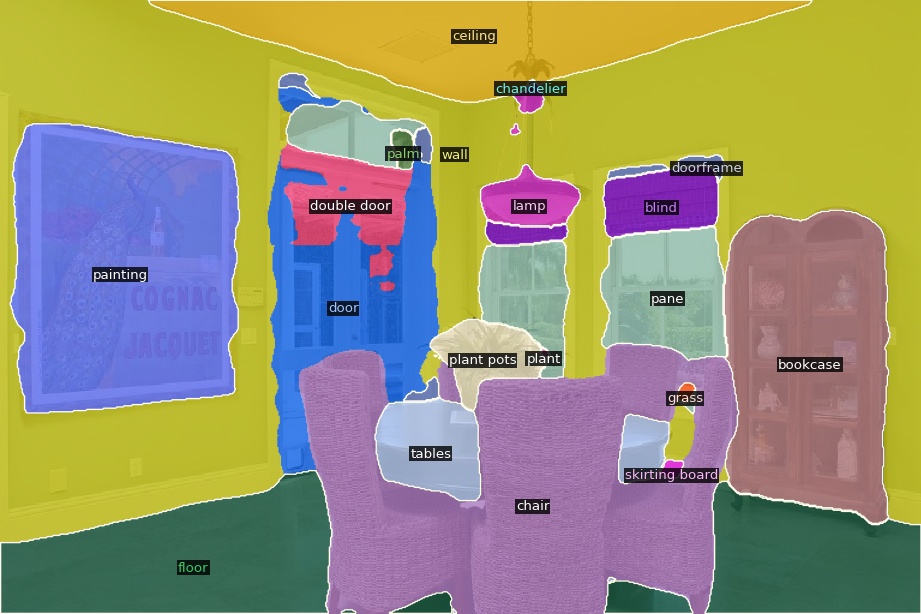} &
\includegraphics[width=0.27\textwidth]{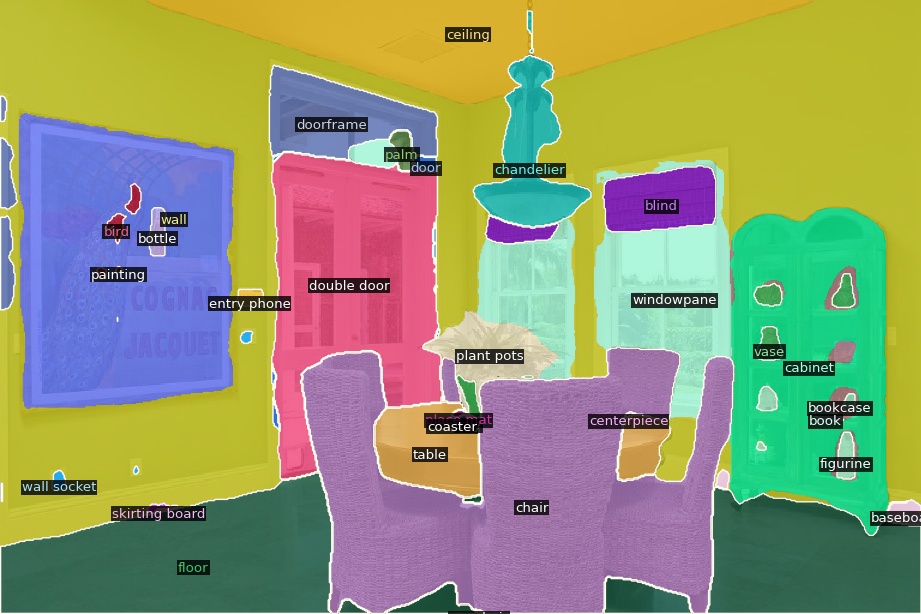} \\

\includegraphics[width=0.27\textwidth]{supple_qual/a847/ADE_val_00001202/image_ADE_val_00001202.jpg} &
\includegraphics[width=0.27\textwidth]{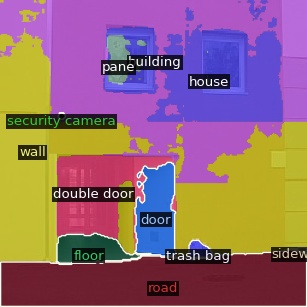} &
\includegraphics[width=0.27\textwidth]{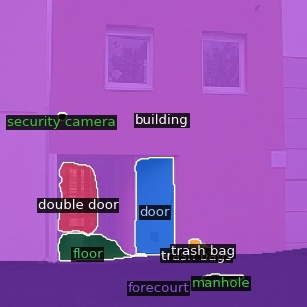} \\

\includegraphics[width=0.27\textwidth]{supple_qual/a847/ADE_val_00001671/image_ADE_val_00001671.jpg} &
\includegraphics[width=0.27\textwidth]{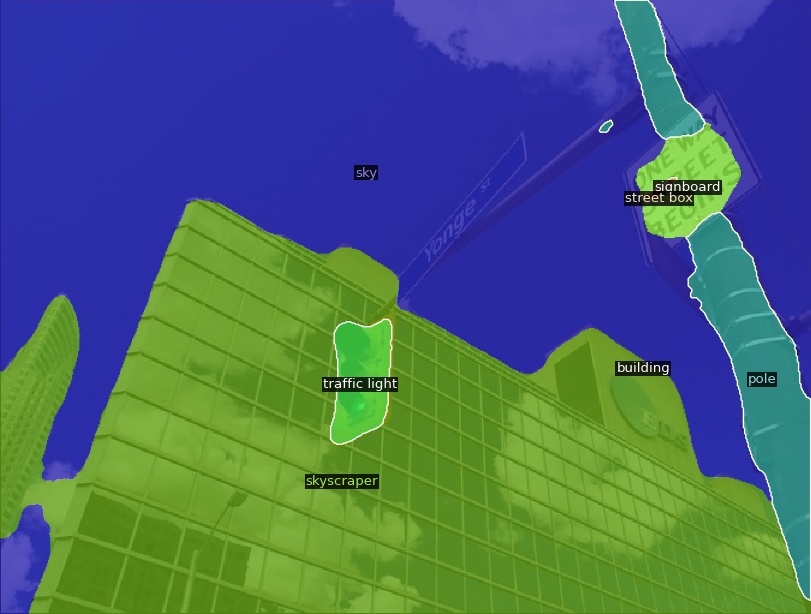} &
\includegraphics[width=0.27\textwidth]{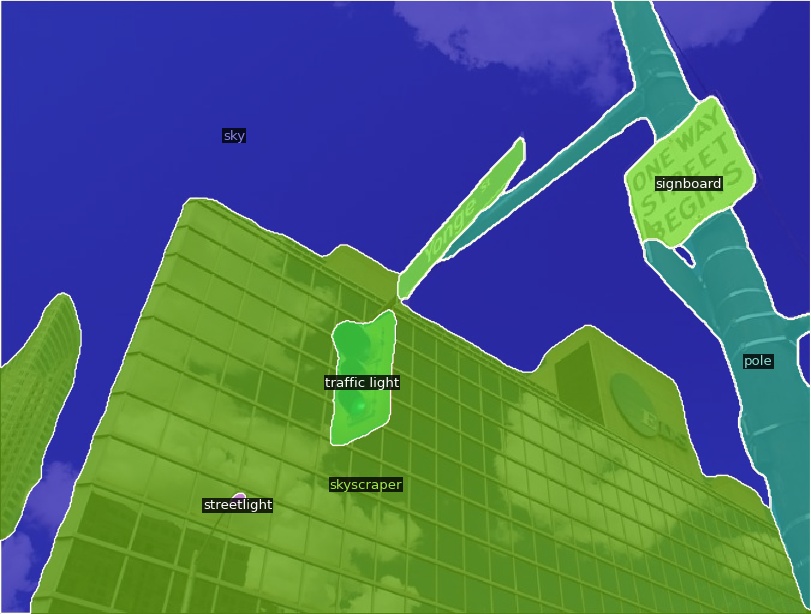} \\

\includegraphics[width=0.27\textwidth]{supple_qual/a847/ADE_val_00000013/image_ADE_val_00000013.jpg} &
\includegraphics[width=0.27\textwidth]{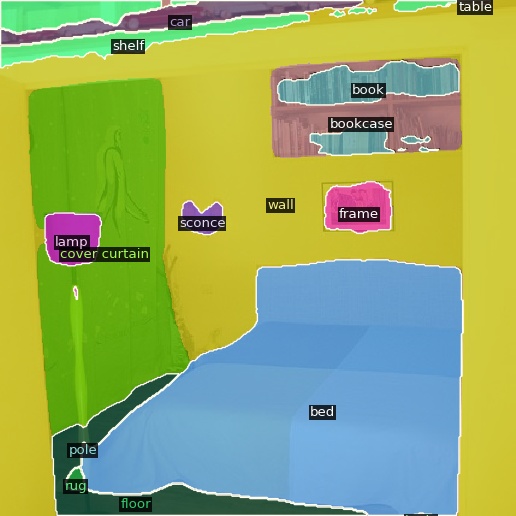} &
\includegraphics[width=0.27\textwidth]{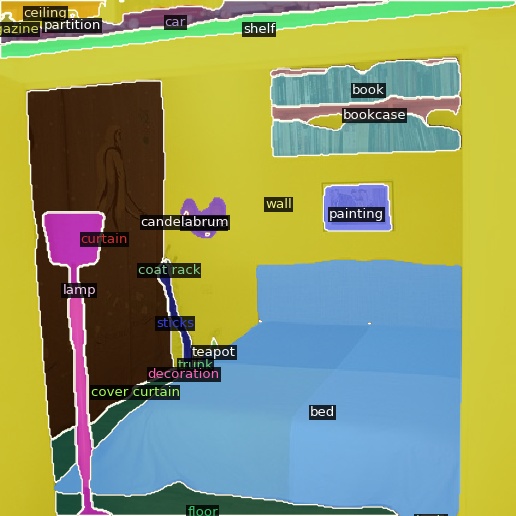} \\

\includegraphics[width=0.27\textwidth]{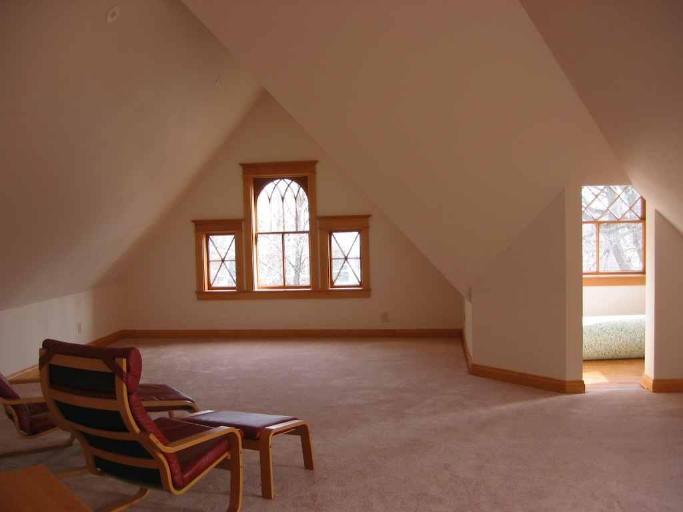} &
\includegraphics[width=0.27\textwidth]{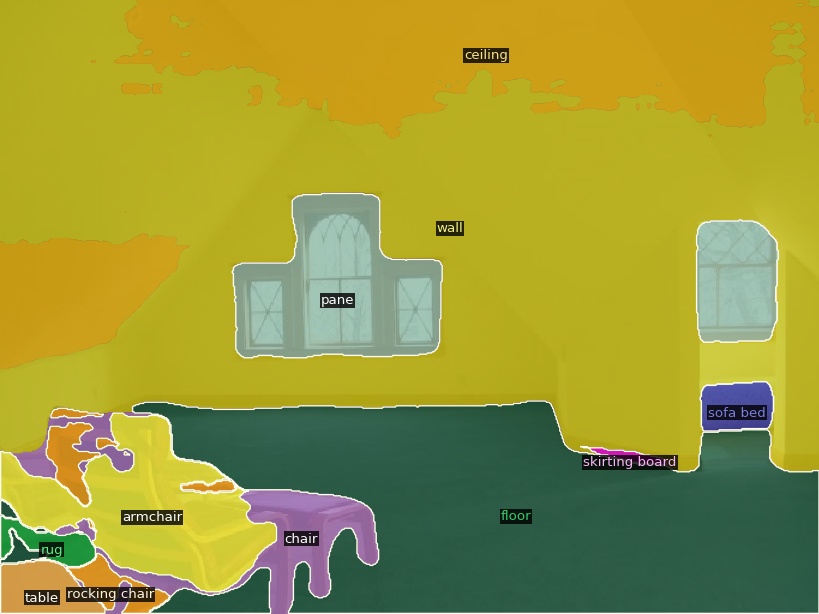} &
\includegraphics[width=0.27\textwidth]{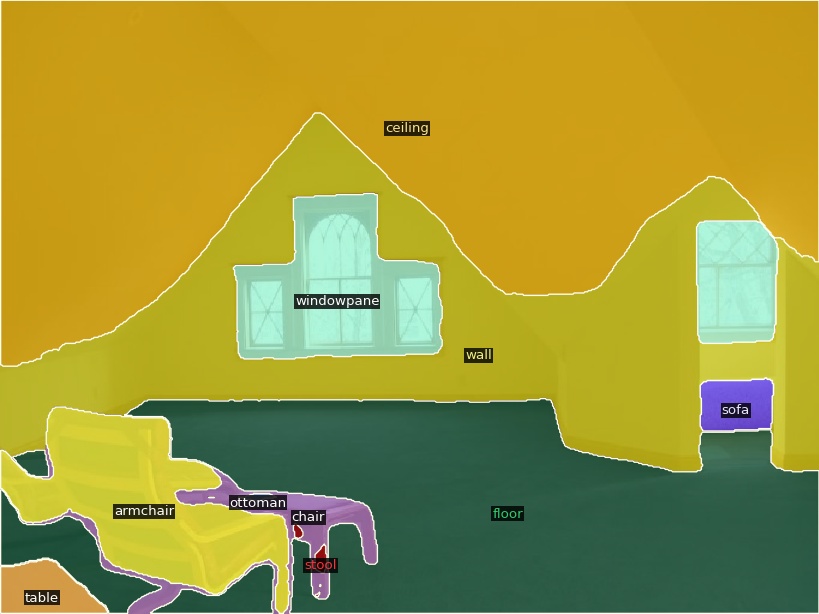} \\

\multicolumn{1}{c}{\small Input Image} &
\multicolumn{1}{c}{\small CAT-Seg} &
\multicolumn{1}{c}{\small Ours} \\[-2pt]

\end{tabular}
\vspace{-5px}
\caption{\textbf{Qualitative results on A-847 dataset.}}
\label{supple_qual_a847}
\end{figure*}

\begin{figure*}[!h]
\centering
\setlength{\tabcolsep}{2pt}
\renewcommand{\arraystretch}{1.0}
\begin{tabular}{ccc}

\includegraphics[width=0.25\textwidth]{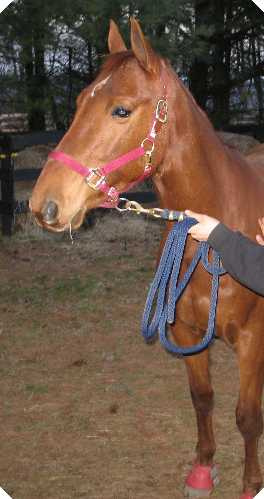} &
\includegraphics[width=0.25\textwidth]{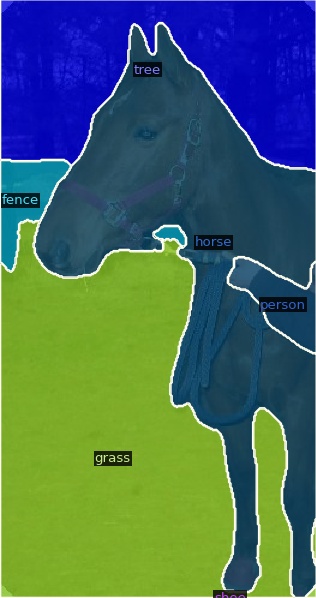} &
\includegraphics[width=0.25\textwidth]{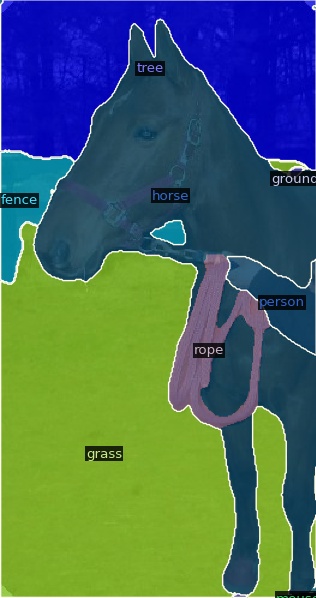} \\

\includegraphics[width=0.25\textwidth]{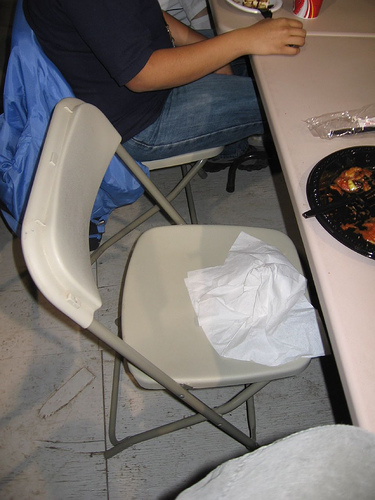} &
\includegraphics[width=0.25\textwidth]{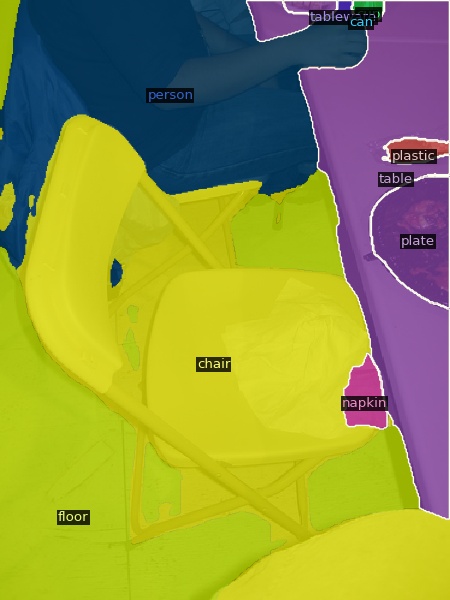} &
\includegraphics[width=0.25\textwidth]{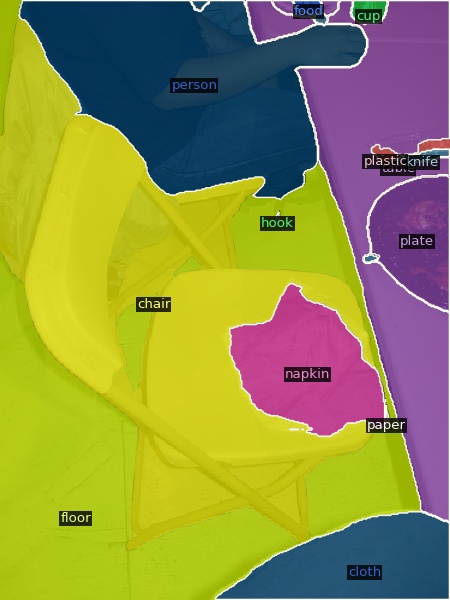} \\

\includegraphics[width=0.25\textwidth]{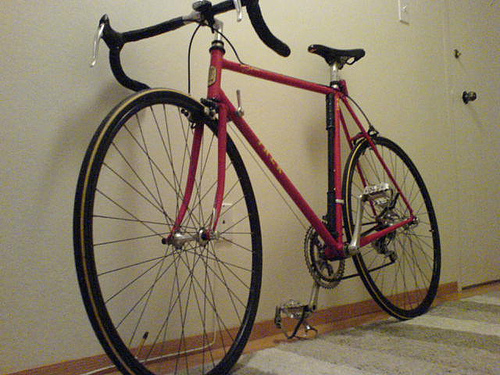} &
\includegraphics[width=0.25\textwidth]{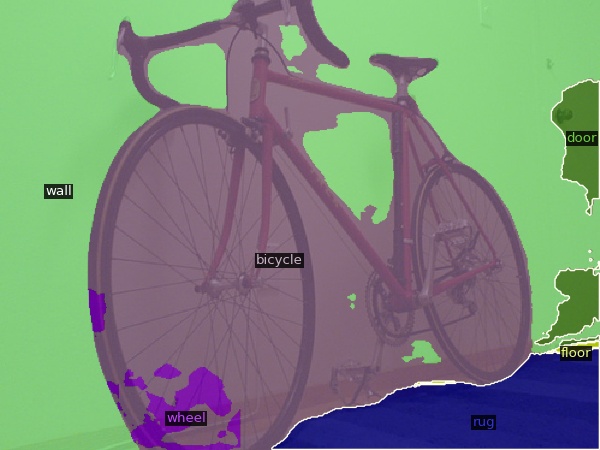} &
\includegraphics[width=0.25\textwidth]{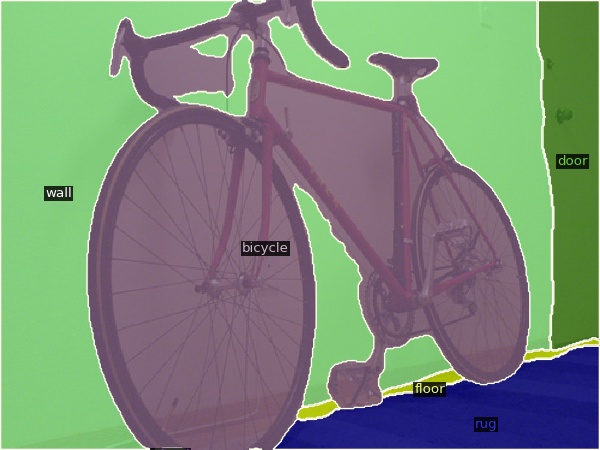} \\

\includegraphics[width=0.25\textwidth]{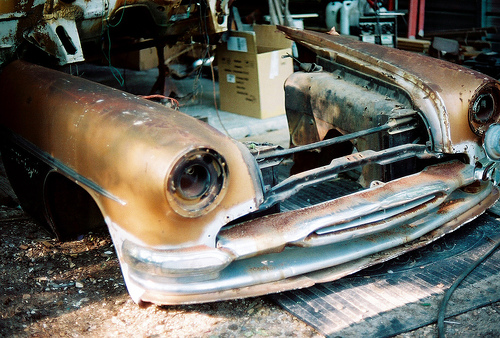} &
\includegraphics[width=0.25\textwidth]{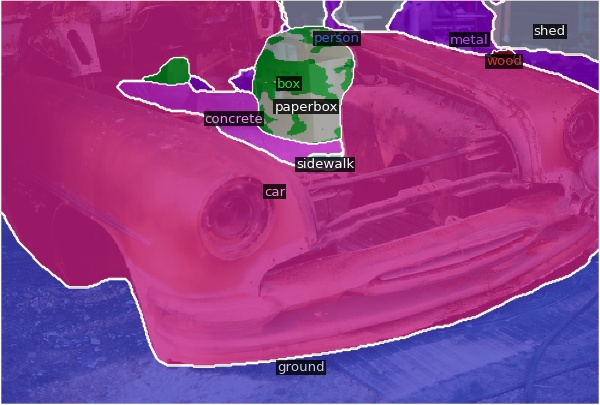} &
\includegraphics[width=0.25\textwidth]{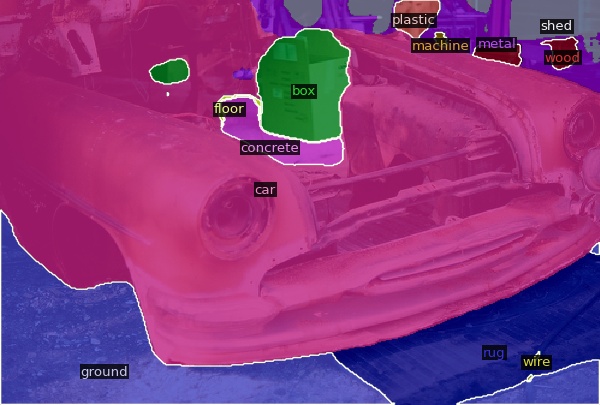} \\

\includegraphics[width=0.25\textwidth]{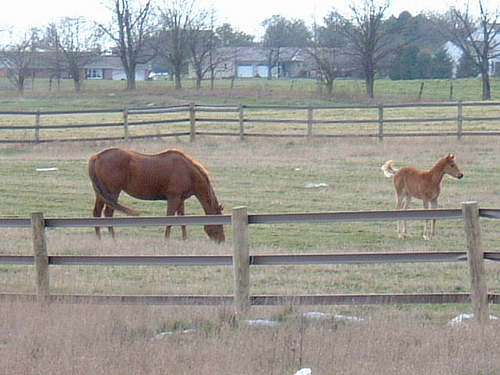} &
\includegraphics[width=0.25\textwidth]{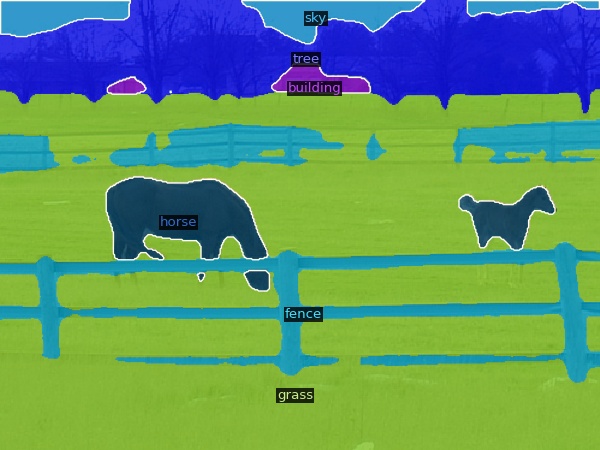} &
\includegraphics[width=0.25\textwidth]{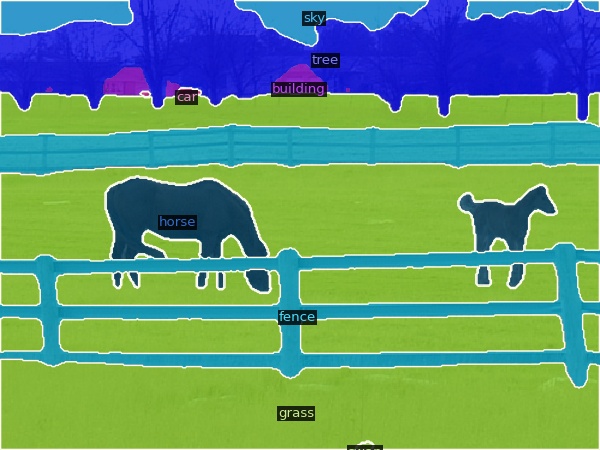} \\

\multicolumn{1}{c}{\small Input Image} &
\multicolumn{1}{c}{\small CAT-Seg} &
\multicolumn{1}{c}{\small Ours} \\[-2pt]

\end{tabular}
\caption{\textbf{Qualitative results on PC-459 dataset.}}
\label{supple_qual_pc459}
\end{figure*}

\begin{figure*}[!h]
\centering
\setlength{\tabcolsep}{2pt}
\renewcommand{\arraystretch}{1.0}
\begin{tabular}{ccc}

\includegraphics[width=0.31\textwidth]{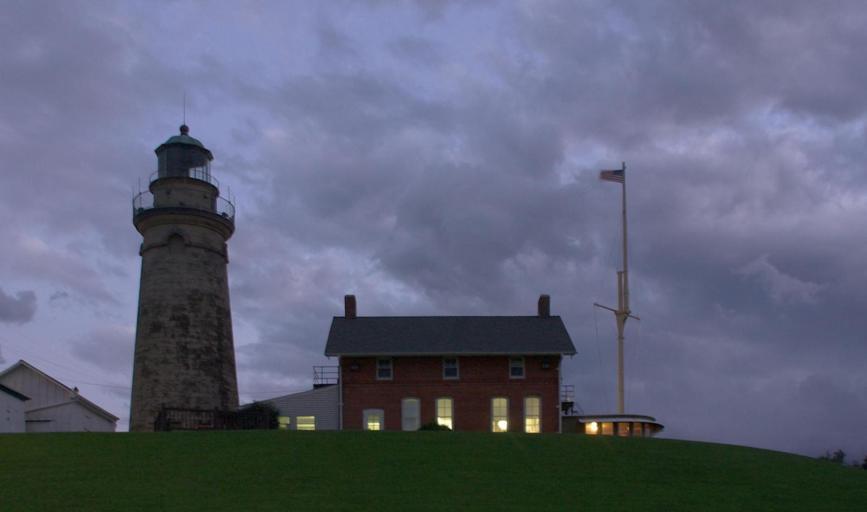} &
\includegraphics[width=0.31\textwidth]{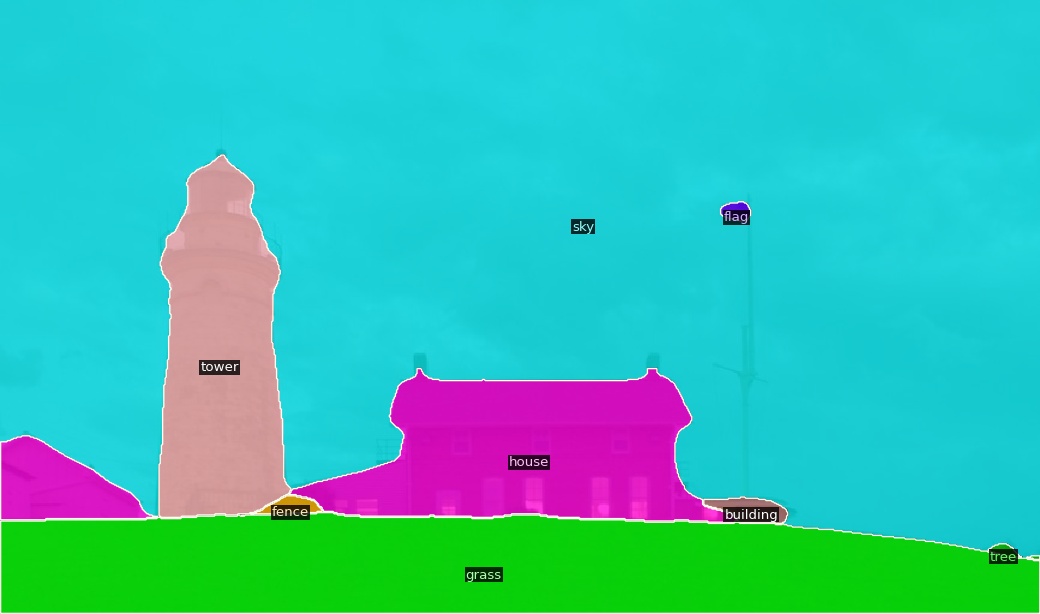} &
\includegraphics[width=0.31\textwidth]{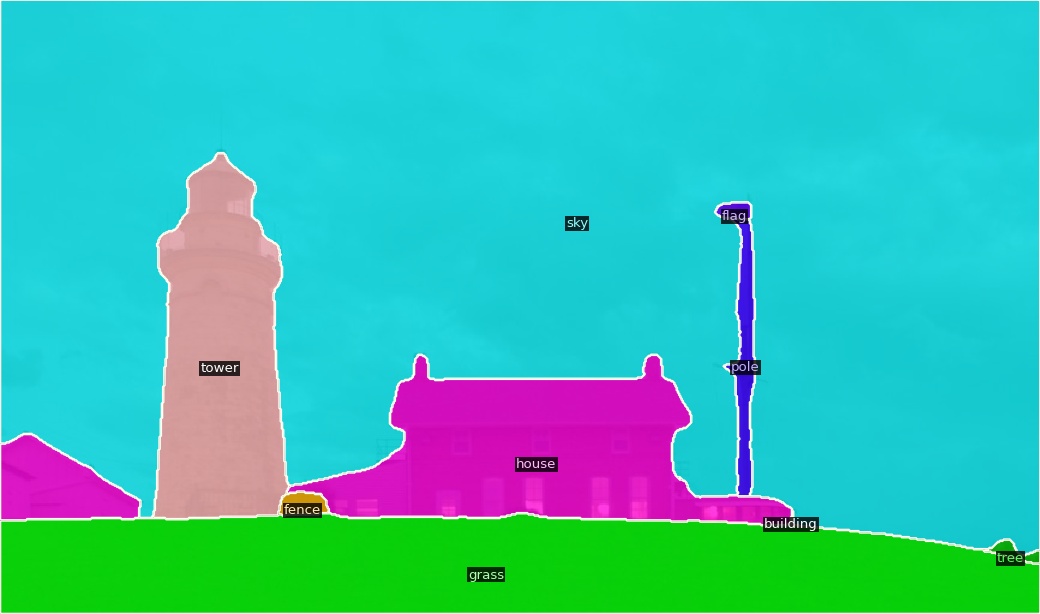} \\

\includegraphics[width=0.31\textwidth]{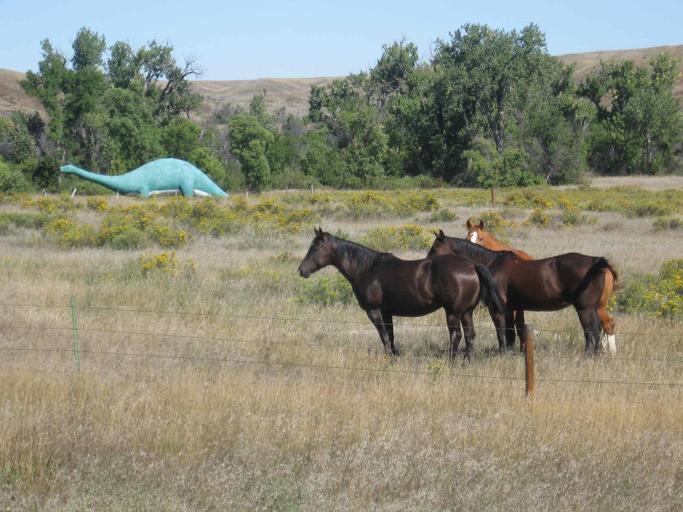} &
\includegraphics[width=0.31\textwidth]{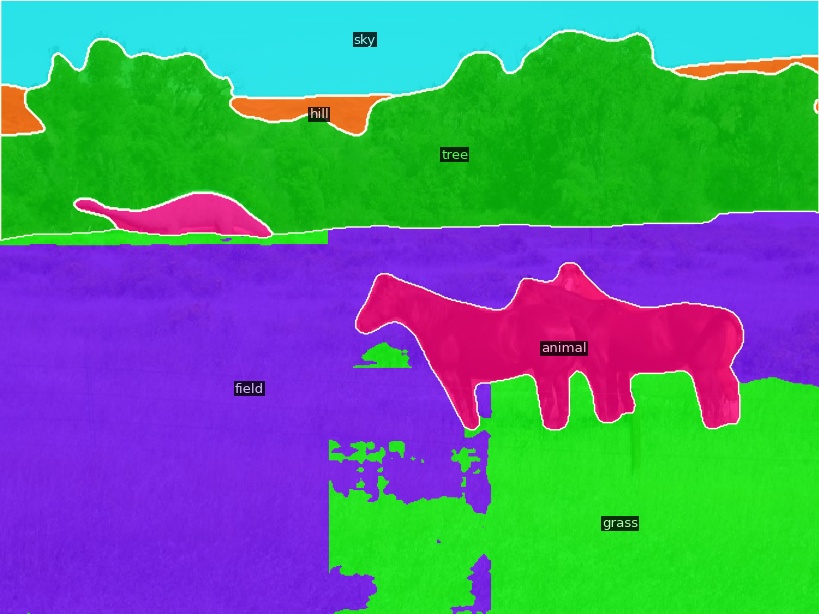} &
\includegraphics[width=0.31\textwidth]{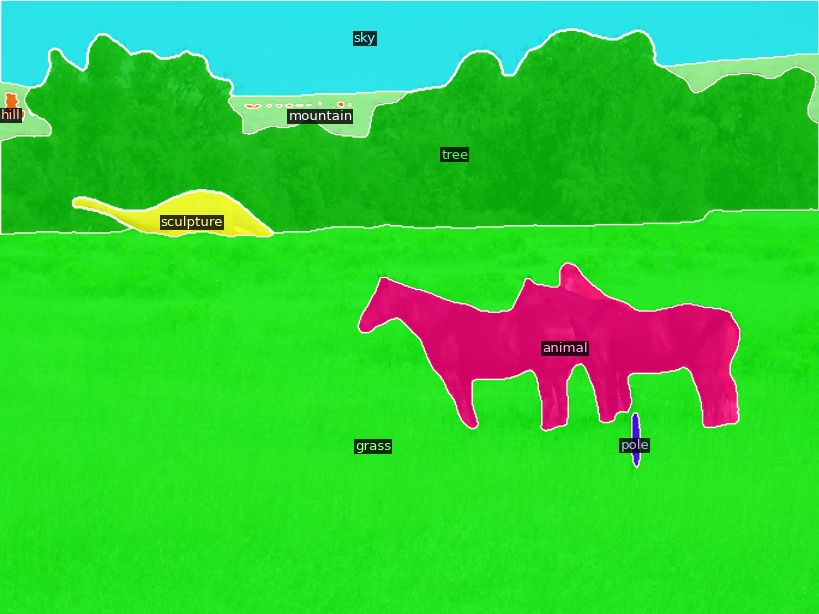} \\

\includegraphics[width=0.31\textwidth]{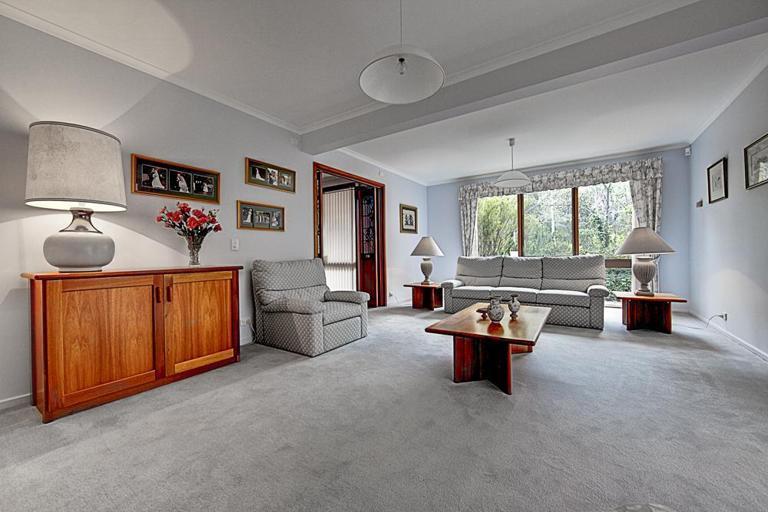} &
\includegraphics[width=0.31\textwidth]{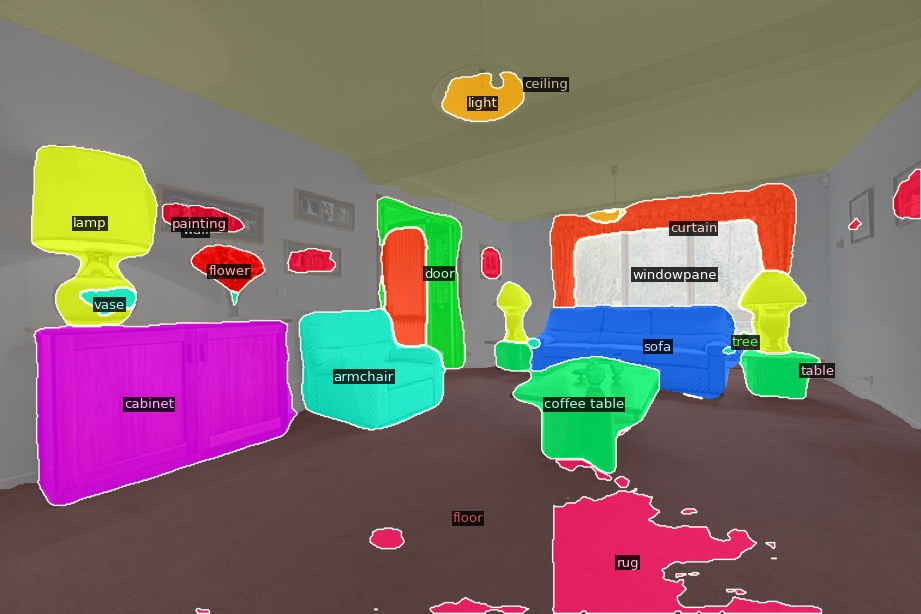} &
\includegraphics[width=0.31\textwidth]{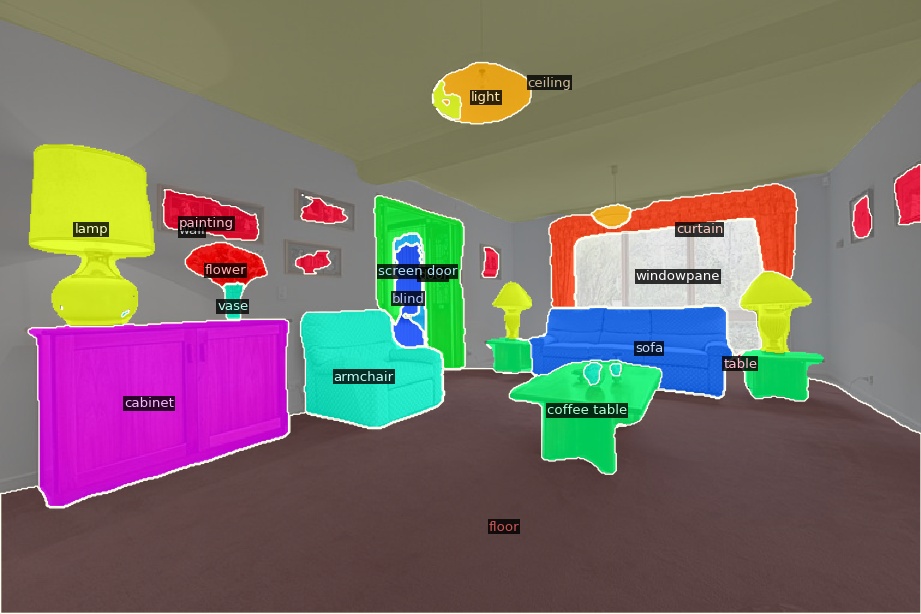} \\

\includegraphics[width=0.31\textwidth]{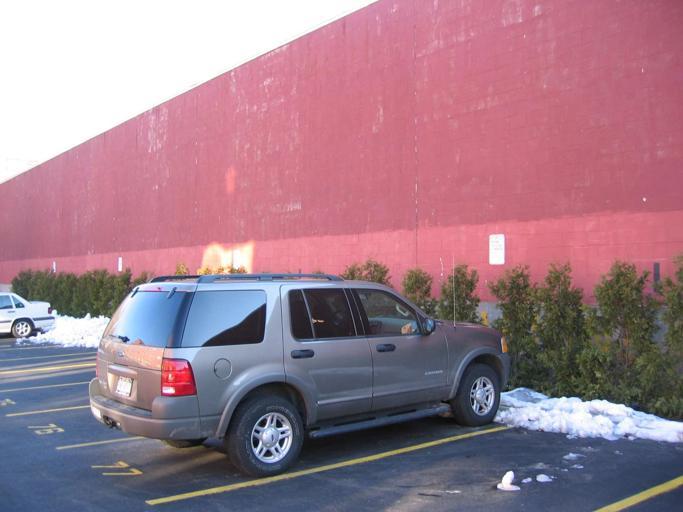} &
\includegraphics[width=0.31\textwidth]{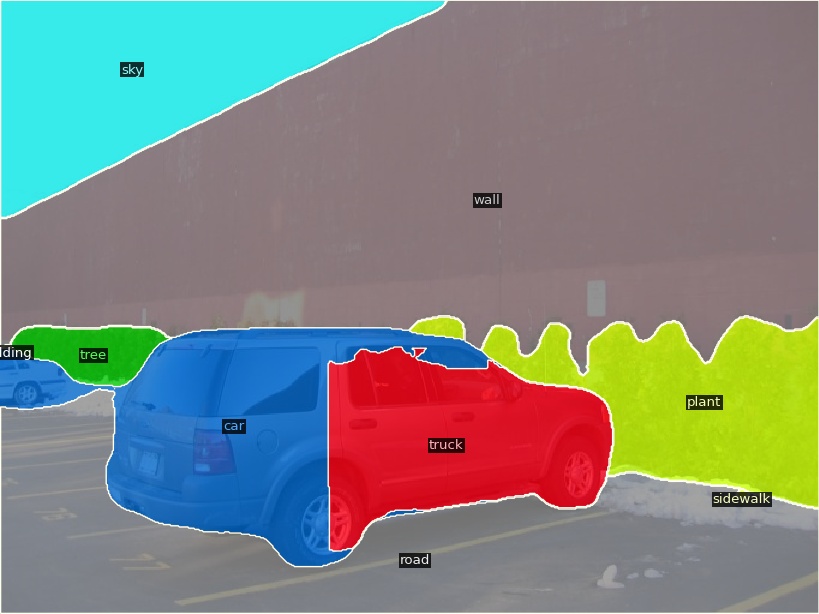} &
\includegraphics[width=0.31\textwidth]{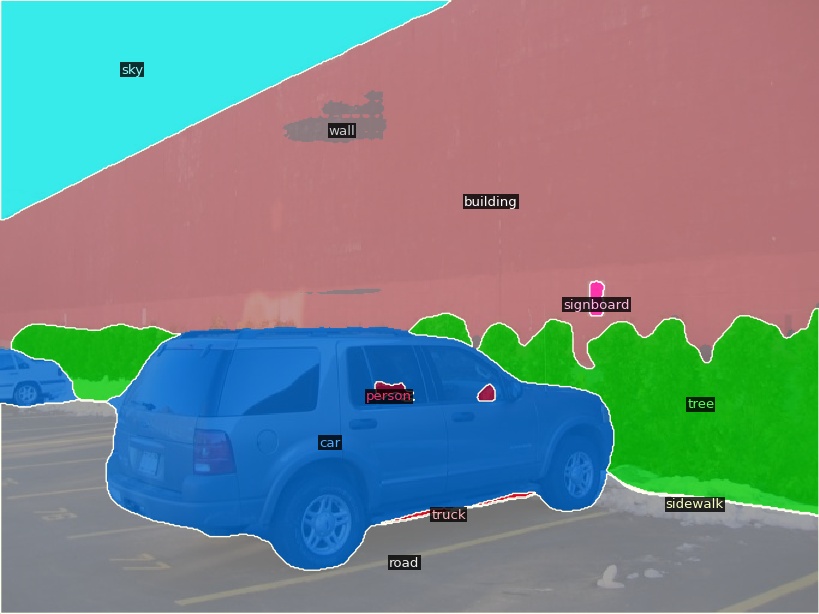} \\

\includegraphics[width=0.31\textwidth]{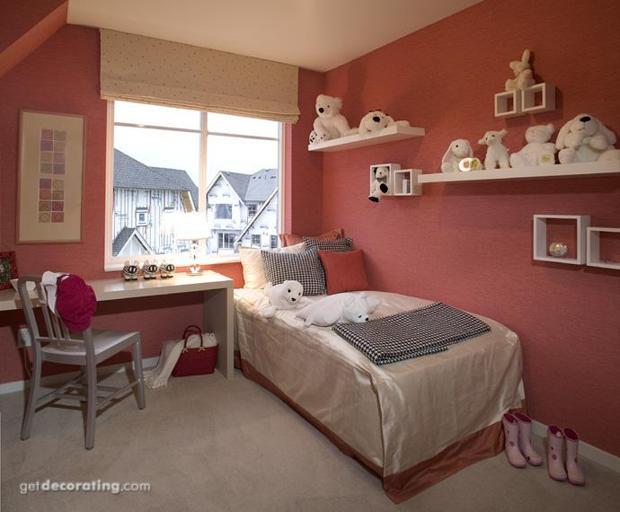} &
\includegraphics[width=0.31\textwidth]{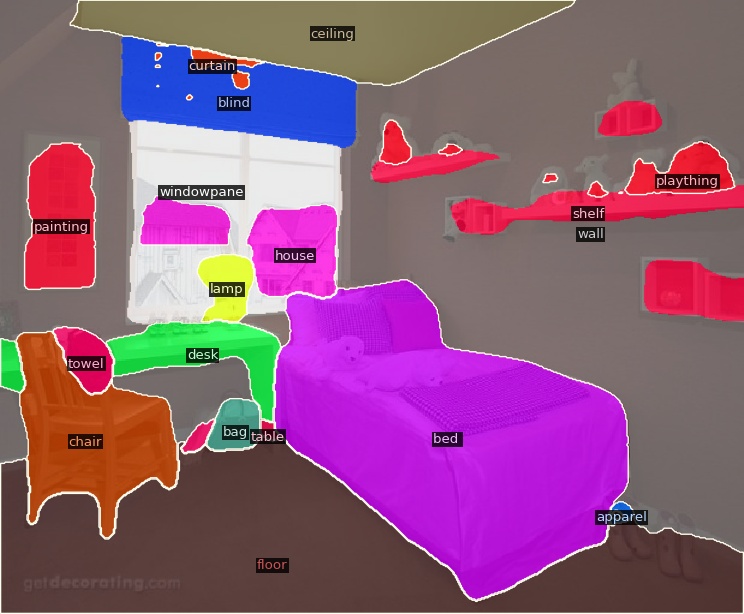} &
\includegraphics[width=0.31\textwidth]{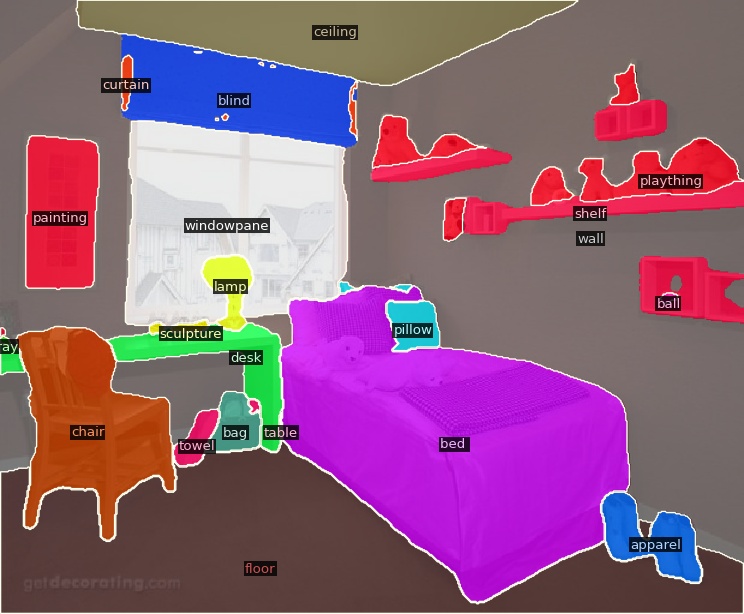} \\

\includegraphics[width=0.31\textwidth]{supple_qual/a150/ADE_val_00001429/image_ADE_val_00001429.jpg} &
\includegraphics[width=0.31\textwidth]{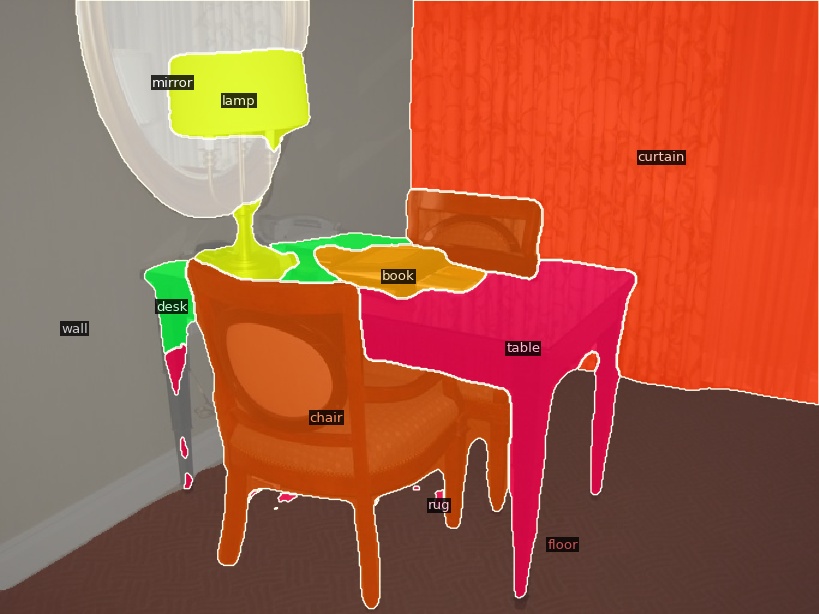} &
\includegraphics[width=0.31\textwidth]{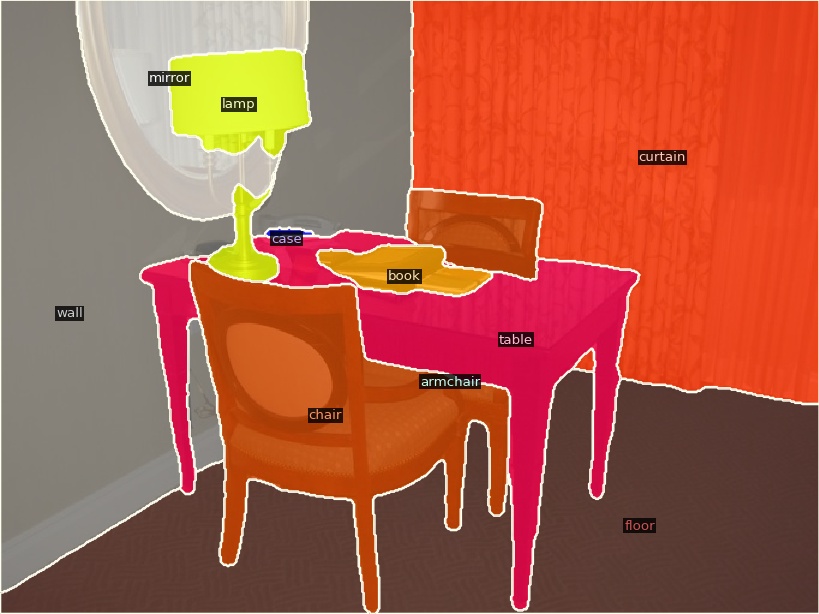} \\

\multicolumn{1}{c}{\small Input Image} &
\multicolumn{1}{c}{\small CAT-Seg} &
\multicolumn{1}{c}{\small Ours} \\[-2pt]

\end{tabular}
\caption{\textbf{Qualitative results on A-150 dataset.}}
\label{supple_qual_a150}
\end{figure*}

\begin{figure*}[!h]
\centering
\setlength{\tabcolsep}{2pt}
\renewcommand{\arraystretch}{1.0}
\begin{tabular}{ccc}

\includegraphics[width=0.30\textwidth]{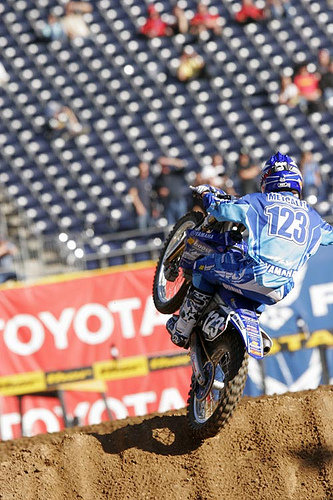} &
\includegraphics[width=0.30\textwidth]{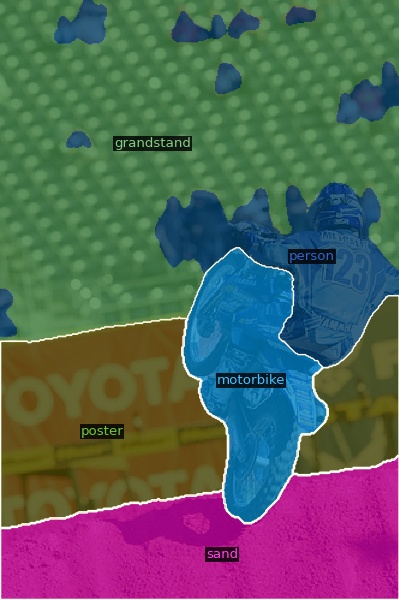} &
\includegraphics[width=0.30\textwidth]{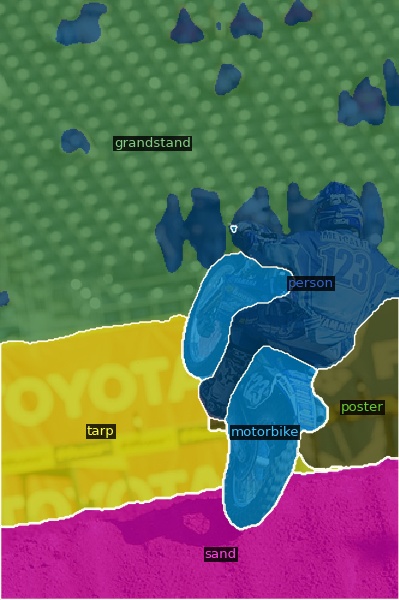} \\

\includegraphics[width=0.30\textwidth]{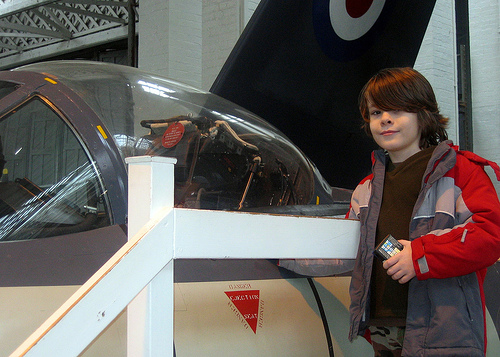} &
\includegraphics[width=0.30\textwidth]{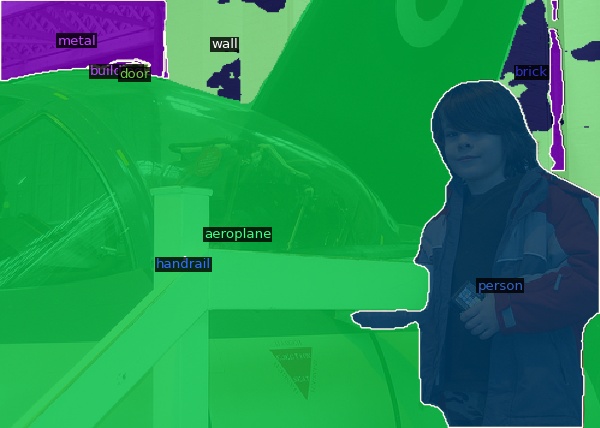} &
\includegraphics[width=0.30\textwidth]{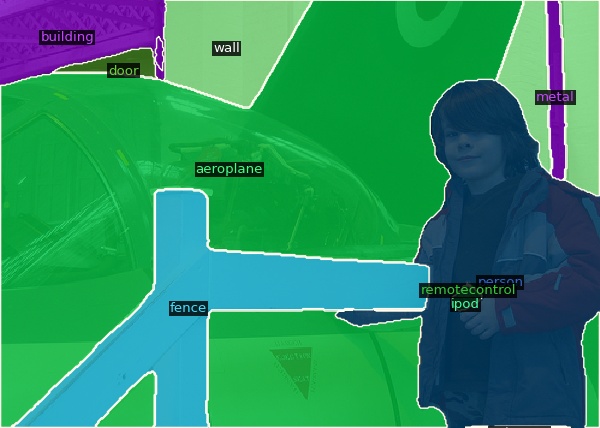} \\

\includegraphics[width=0.30\textwidth]{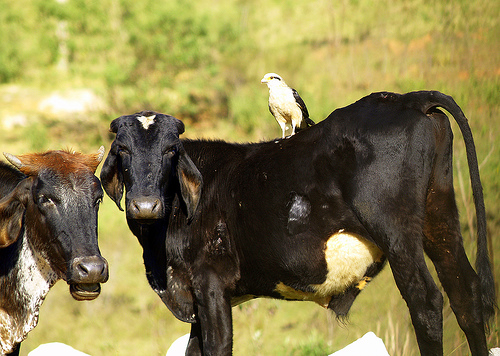} &
\includegraphics[width=0.30\textwidth]{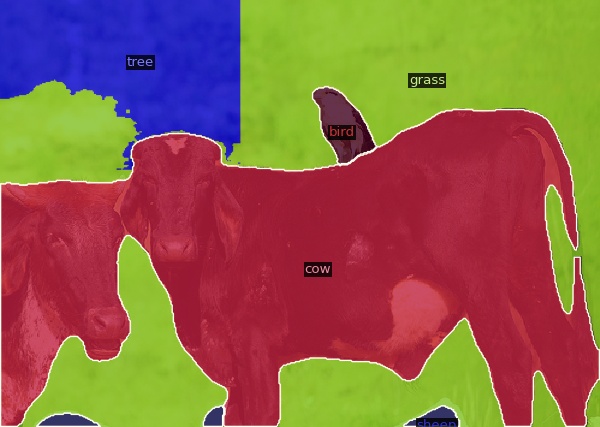} &
\includegraphics[width=0.30\textwidth]{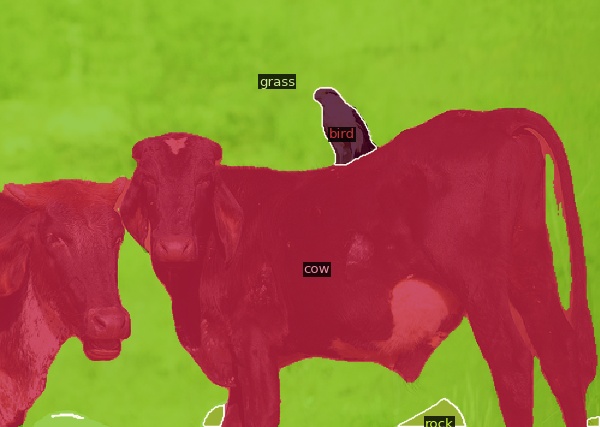} \\

\includegraphics[width=0.30\textwidth]{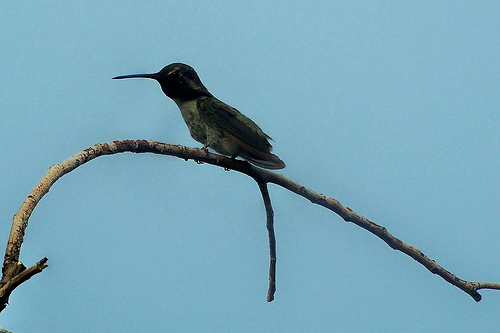} &
\includegraphics[width=0.30\textwidth]{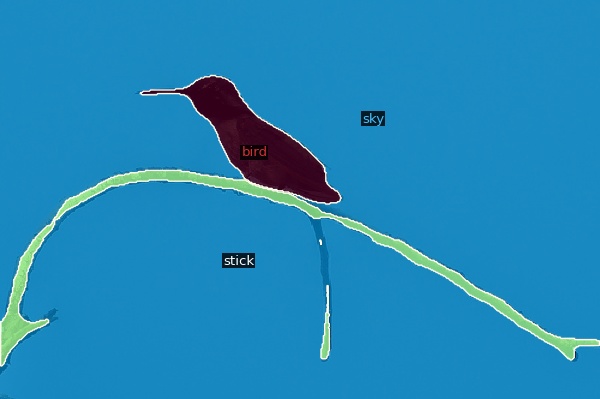} &
\includegraphics[width=0.30\textwidth]{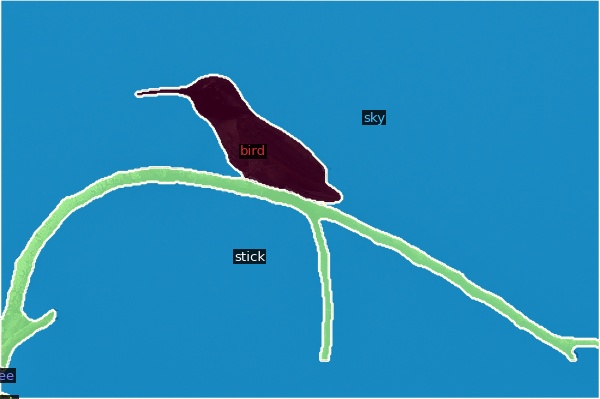} \\

\multicolumn{1}{c}{\small Input Image} &
\multicolumn{1}{c}{\small CAT-Seg} &
\multicolumn{1}{c}{\small Ours} \\[-2pt]

\end{tabular}
\caption{\textbf{Qualitative results on PC-59 dataset.}}
\label{supple_qual_pc59}
\end{figure*}

\vspace{-10px}
We show qualitative comparison with CAT-Seg on A-847, PC-459, A-150 and PC-59 datasets in Fig.~\ref{supple_qual_a847},~\ref{supple_qual_pc459},~\ref{supple_qual_a150},~\ref{supple_qual_pc59} respectively.

\end{document}